\documentclass{article}

\usepackage{arxiv}

\usepackage[utf8]{inputenc} 
\usepackage[T1]{fontenc}    
\usepackage{lmodern}        
\usepackage{hyperref}       
\usepackage{url}            
\usepackage{booktabs}       
\usepackage{amsfonts}       
\usepackage{nicefrac}       
\usepackage{microtype}      
\usepackage{graphicx}

\title{Pretty darn good control: when are approximate solutions better
than approximate models}

\author{
    Felipe Montealegre-Mora
   \\
     \\
   \\
  \texttt{} \\
   \And
    Marcus Lapeyrolerie
   \\
     \\
   \\
  \texttt{} \\
   \And
    Melissa Chapman
   \\
     \\
   \\
  \texttt{} \\
   \And
    Abigail G. Keller
   \\
     \\
   \\
  \texttt{} \\
   \And
    Carl Boettiger
   \\
     \\
   \\
  \texttt{\href{mailto:cboettig@berkeley.edu}{\nolinkurl{cboettig@berkeley.edu}}} \\
  }

\newlength{\cslhangindent}
\newlength{\csllabelwidth}
\newlength{\cslentryspacingunit} 
  {}%
  {\par}
\newenvironment{CSLReferences}[2] 
 {
  \setlength{\parindent}{0pt}
  \ifodd #1
  \let\oldpar\par
  \def\par{\hangindent=\cslhangindent\oldpar}
  \fi
  \setlength{\parskip}{#2\cslentryspacingunit}
 }%
 {}
\usepackage{calc}

\usepackage{amsmath}
\usepackage{amssymb}
\usepackage{setspace}
\usepackage{bookmark}
\usepackage{float}
\usepackage{graphicx}
\usepackage{caption}
\usepackage{subcaption}
\usepackage{tikz-cd}
\usepackage{tcolorbox}
\usepackage{etoolbox}
\usepackage{url}

\begin{document}
\maketitle

\begin{abstract}
Existing methods for optimal control struggle to deal with the
complexity commonly encountered in real-world systems, including
dimensionality, process error, model bias and data heterogeneity.
Instead of tackling these system complexities directly, researchers have
typically sought to simplify models to fit optimal control methods. But
when is the optimal solution to an approximate, stylized model better
than an approximate solution to a more accurate model? While this
question has largely gone unanswered owing to the difficulty of finding
even approximate solutions for complex models, recent algorithmic and
computational advances in deep reinforcement learning (DRL) might
finally allow us to address these questions. DRL methods have to date
been applied primarily in the context of games or robotic mechanics,
which operate under precisely known rules. Here, we demonstrate the
ability for DRL algorithms using deep neural networks to successfully
approximate solutions (the ``policy function'' or control rule) in a
non-linear three-variable model for a fishery without knowing or ever
attempting to infer a model for the process itself. We find that the
reinforcement learning agent discovers a policy that outperforms both
constant escapement and constant mortality policies---the standard
family of policies considered in fishery management. This DRL policy has
the shape of a constant escapement policy whose escapement values depend
on the stock sizes of other species in the model.
\end{abstract}

\keywords{
    Optimal Control
   \and
    Reinforcement Learning
   \and
    Uncertainty
   \and
    Decision Theory
  }

\hypertarget{intro}{%
\section{Introduction}\label{intro}}

Much effort has been spent grappling with the complexity of our natural
world in contrast to the relative simplicity of the models we use to
understand it. Heroic amounts of data and computation are being brought
to bear on developing better, more realistic models of our environments
and ecosystems, in hopes of improving our capacity to address the many
planetary crises. But despite these efforts and advances, we remain
faced with the difficult task of figuring out how best to respond to
these crises. While simplified process models for the population
dynamics have historically allowed for exploration of large decision
spaces, the new wave of rich models are applied to highly oversimplified
descriptions of potential actions they seek to inform. For instance,
Global Circulation Models (GCMs) such as HadCM3 (Pope et al. 2000; C.
Gordon et al. 2000; Collins, Tett, and Cooper 2001) model earth's
climate using 1.5M variables, while the comparably vast potential action
space is modeled much more minimalistically, with 5 SSP socioeconomic
storylines and 7 SSP-RCP marker scenarios summarizing the action space
at the IPCC (Riahi et al. 2017).

Even as our research community develops simulations of the natural world
that fit only in supercomputers, we analyze a space of policies that
would fit on index cards. Similar combinations of rich process models
and highly simplified decision models (often not even given the status
of `model') are common. Modeling the potential action space as one of a
handful of discrete scenarios is sometimes a well justified
acknowledgement of the constraints faced by real-world decision-makers
-- particularly in the context of multilateral decisions -- and may seem
to reflect a division of responsibilities between `scientists' modeling
the `natural processes' and policy-makers who make the decisions. But,
more often, this simplification of decision choices is simply
mathematically or conceptually convenient. This simplification reflects
trade-offs between tractablity and complexity at the basis of any
mathematical modeling -- if we make both the state space and action
space too realistic, the problem of finding the best sequence of actions
quickly becomes intractable. However, emerging data-driven methods from
machine learning offer a new choice -- algorithms that can find good
strategies in previously intractable problems, but at the cost of
opacity.

In this paper, we focus on a well-developed application of model-based
management of the natural world that has long illustrated the trade-offs
between model complexity and policy complexity: the management of marine
fisheries. Fisheries management is both an important issue to society as
well as a rich and frequent test-bed of ecological management more
generally. Fisheries are an essential natural resource that provide the
primary source of protein for one in every four humans, and have faced
widely documented declines due to over-fishing Costello et al. (2016).
Fisheries management centers around the process of sampling populations
to determine fishing quotas based on population estimates. This decision
is often guided by a model of the dynamics of the system. Our paper
focuses on the decision side of this problem rather than the measurement
step.

Fisheries management has roots in both the fields of \emph{ecosystem
management} and \emph{natural resource economics}. Both fields might
trace their origins to the notion of maximum sustainable yield (MSY),
introduced independently by a fisheries ecologist (Schaefer 1954) and
the economist (H. S. Gordon and Press 1954) in the same year. From this
shared origin, each field would depart from the simplifying assumptions
of the Gordon-Schaefer model in divergent ways, leading to different
techniques for deriving policies from models.The heart of the management
problem is easily understood: a manager seeks to set quotas on fishing
that will ensure the long-term profitability and sustainability of the
industry. Mathematical approaches developed over the past century may be
roughly divided between these two fields: (A) ecologists, focused on
ever more realistic models of the biological processes of growth and
recruitment of fish while considering a relatively stylized suite of
potential management strategies, and (B) economists, focused on far more
stylized models of the ecology while exploring a far less constrained
set of possible policies. The economist's approach can be characterized
by the mathematics of a Markov decision process (MDP Colin W. Clark
1973; Colin W. Clark 1990; Marescot et al. 2013), in which the
decision-maker must observe the stock each year and recommend a possible
action. In this approach, the policy space that must be searched is
exponentially large -- for a management horizon of T decisions and a
space of N actions, the number of possible policies is \(N^T\). In
contrast, fisheries ecologists and ecosystem management typically search
a space of policies that does not scale with the time horizon. Under
methods such as ``Management Strategy Evaluation'' (MSE, (Punt et al.
2016)) a manager identifies a candidate set of ``strategies'' a priori,
and then compares the performance of each strategy over a suite of
simulations to determine which strategy gives the best outcome
(i.e.~best expected utility). This approach is far more amenable to
complex simulations of fisheries dynamics and more closely corresponds
to how most marine fishing quotas are managed today in the United States
(see stock assessments documented in RAM Legacy Stock Assessment
Database 2020).

\includegraphics[width=1\linewidth]{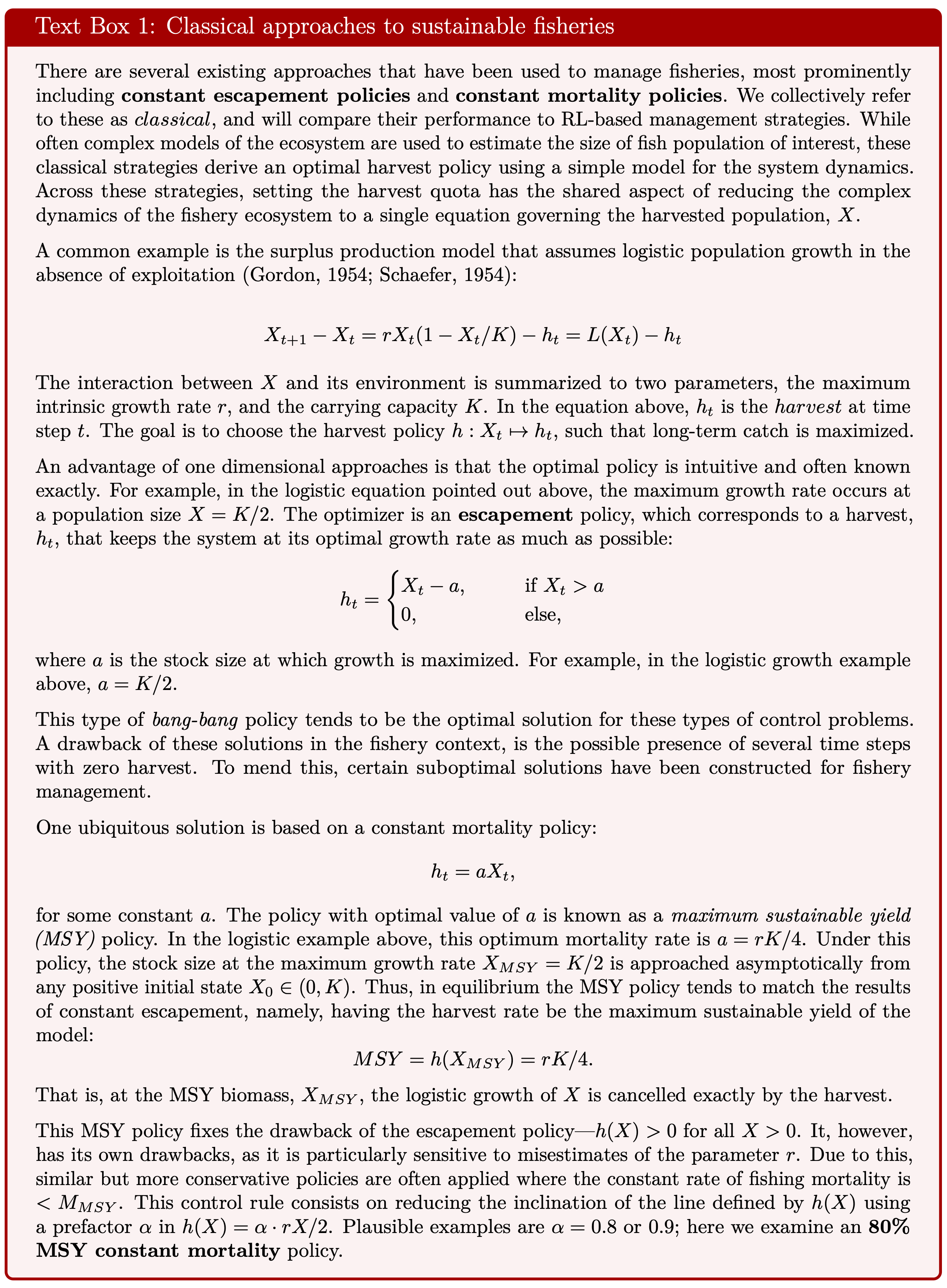}

Recent advances in machine learning may allow us to once again bridge
these approaches, while also bringing new challenges of their own. Novel
data-driven methods have allowed these models to evolve into ever more
complex and realistic simulations used in fisheries management, where
models with over 100 parameters are not uncommon (RAM Legacy Stock
Assessment Database 2020). Constrained by computational limits, MDP
approaches have been intractable on suitably realistic models and
largely confined to more academic applications (Costello et al. 2016).
However, advances in \emph{Deep Reinforcement Learning}, (DRL) a
sub-field of machine learning, have recently demonstrated remarkable
performance in a range of such MDP problems, from video games (Bellemare
et al. 2013; Mnih et al. 2013) to fusion reactions (Degrave et al. 2022;
Seo et al. 2022) to the remarkable dialog abilities of ChatGPT (OpenAI
2022). RL methods also bring many challenges of their own: being
notoriously difficult to train and evaluate, requiring immense
computational costs, and presenting frequent challenges with
reproducibility. A review of all these issues is beyond our scope but
can be found elsewhere (Lapeyrolerie et al. 2022; Chapman et al. 2023).
Here, though, we will focus on the issue of opacity and interpretability
raised by these methods. In contrast with optimization algorithms
currently used in either ecosystem management or resource economics, RL
algorithms have no guarantees of or metrics for convergence to an
optimal solution. In general, one can only assess the performance of
these black box methods relative to alternatives.

In most US fisheries, the mortality policy is often piecewise linear
(often with one constant and one linear piece), and the allowable
biological catch (ABC) or total allowable catch (TAC) is set at some
heuristic (e.g.~80\%) below the `overfishing limit', \(F_{MSY}\),
i.e.~the highest (constant) mortality that can be sustained indefinitely
(in the model -- reality of course does not permit such definitions).
This fixed mortality management can be seen, for instance, in most of
the fisheries listed in the widely used R.A. Myers Legacy Stock
Assessment Database. Here, we have focused on purely constant mortality
policies, rather than piecewise linear mortality funcitons, for
simplicity. Escapement-based management is less common, except in
salmonoids, as it requires closing a fishery whenever the measured
biomass falls below \(B_{MSY}\).

In this article, we compare against two common methods: constant
mortality (CMort) and constant escapement (CEsc), introduced in Text Box
1.\footnote{
A repository with all the relevant code to reproduce our results may be found at \url{https://github.com/boettiger-lab/approx-model-or-approx-soln} in the ‘‘src’’ subdirectory. 
The data used is found in the ‘‘data’’ subdirectory, but the user may use the code provided to generate new data sets.
}

We consider the problem of devising harvest strategies in a for a series
of ecosystem models with increasing complexity (Table \ref{tab:models}):
\emph{1) one species, one fishery:} a simple single-species recruitment
model based on (May 1977); \emph{2) three species, one fishery:} a
three-species generalization of model \emph{1)}, where one of the
species is harvested; \emph{3) three species, two fisheries:} the same
three-species model as above but with two harvested species; \emph{4)
three species, two fisheries, parameter variation:} a three-species
model of which two are harvested, as above, with a time-varying
parameter. This last model is meant to be a toy model of climate
change's effect on the system. Across all of these scenarios, two goals
are balanced in the decision process: maximizing long-term catch and
preventing stock sizes to fall below some a-priori threshold.

\begin{table}
  \begin{center}
    \begin{tabular}{|l|c|c|c|c|}
      \hline
      \textbf{Model name} & \textbf{Model eqs.} & \textbf{N. Sp.} & \textbf{Harv. Sp.} & \textbf{Stationary?}\\
      \hline
      Model 1 & \eqref{eq:may} & 1 & $X$ & Yes \\
      Model 2 & \eqref{eq:3d model} & 3 & $X$ & Yes \\
      Model 3 & \eqref{eq:3d model} & 3 & $X$ and $Y$ & Yes \\
      Model 4 & \eqref{eq:3d model} & 3 & $X$ and $Y$ & No \\
      \hline
    \end{tabular}
    \caption{Table of models considered in this paper. Here, \textbf{N. Sp.} is the \emph{number of species} in the model, \textbf{Harv. Sp.} is the species of the model which are harvested, and \textbf{Stationary?} refers to whether the parameters of the model have fixed values (or, on the contrary, if they vary in time). The only non-stationary case presented in the paper is where $r_X$ drifts linearly with time. In the code repository associated to the paper, we consider other possible choices of non-stationarity.}
    \label{tab:models}
  \end{center}
\end{table}

\begin{figure}[H]

{\centering \includegraphics[width=6.5in]{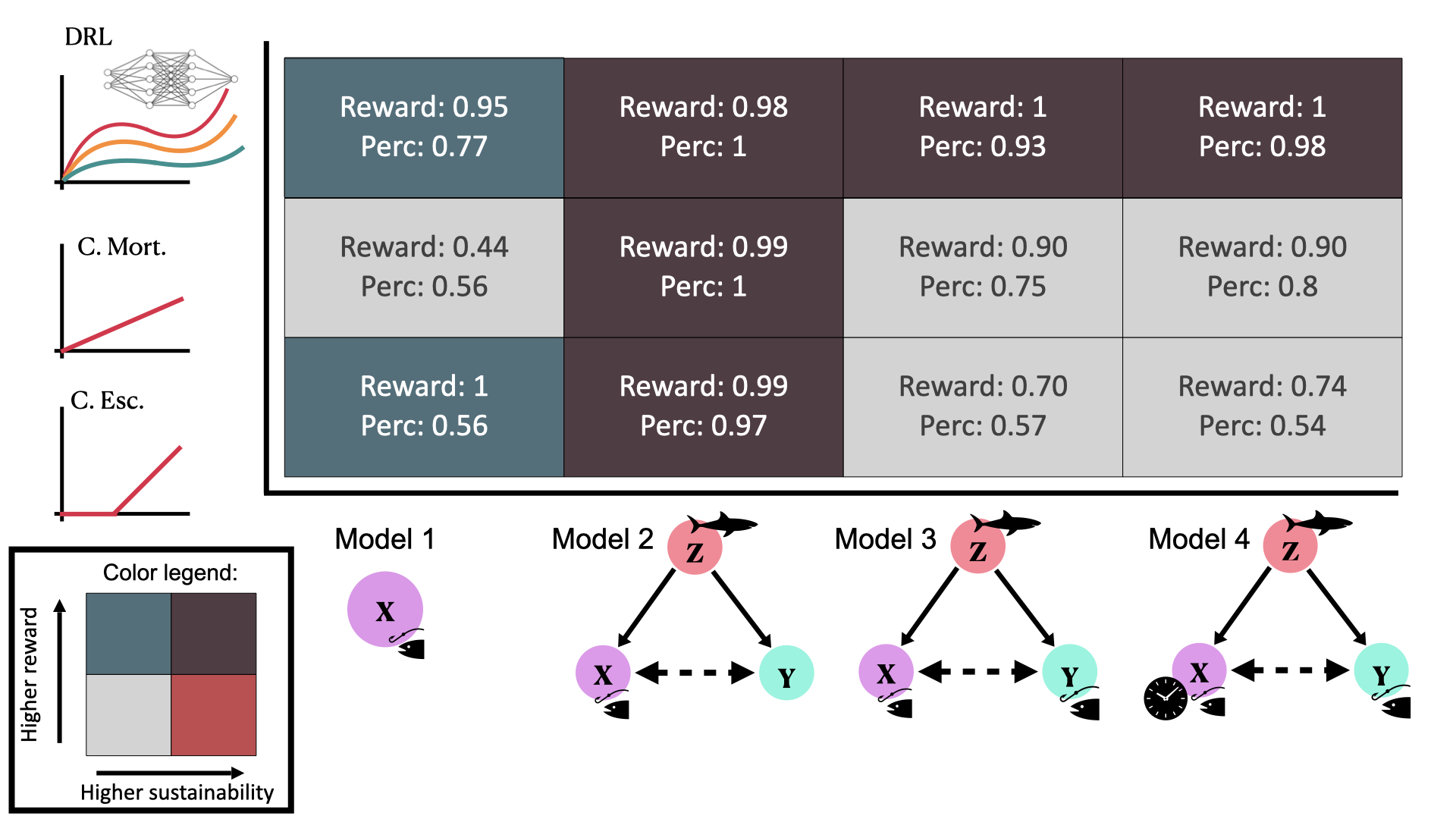} 

}

\caption{An experimental-design type of visualization of the management scenarios considered in this paper. On the x-axis are four different fishery management problems (Table 1). We represent the Model 4’s non-stationarity with a clock next to the X variable, and we intend to use it as an example of a possible simplified model for the effects of climate change. On the y-axis we have different management strategies with which one may control each of the models. On the bottom we have the constant escapement strategy (CEsc), based on calling off all fishing below a certain threshold population value. Above that is the constant mortality strategy (CMort), where one optimizes over constant fishing effort strategies. Finally, on top we have DRL-based strategies where policies are in general functions of the full state of the system, and they are parametrized by a neural network. The specific DRL-based strategy is referred to as PPO+GP in the main text, due to the algorithm used to produce the policy. The results plotted are the average reward obtained by the strategy over 100 episodes, and the fraction of those episodes which do not end with a near-extinction event (denoted Perc for Percentage). We have normalized to the highest reward in each column in order to enhance the comparison between strategies. For illustrative purposes we have color-coded the results using a two-dimensional color legend displayed on the bottom left.}\label{fig:conceptual-v2}
\end{figure}

This way, we evaluate classical management strategies (CMort and CEsc)
and DRL-based strategies on four different models. This experimental
design is summrized in Fig. \ref{fig:conceptual-v2}.

Regarding control for these 4 models, we show the following. \emph{Model
1:} DRL-based strategies recover the optimal constant escapement (CEsc)
policy function. Constant mortality (CMort) performs considerably worse
than these three.\footnote{
As will be explained later, all our models are stochastic. 
If we set stochasticity to zero in Model 1, CMort matches the performance of the other management strategies.
} \emph{Model 2:} Here, all management strategies perform similarly.
\emph{Model 3:} For this model, DRL outperforms both classical
strategies, with CMort surprisingly performing significantly better than
CEsc. In particular, we observe that DRL strategies are more sensitive
to stochastic variations of the system which allows it to adaptively
manage the system to prevent population collapses below the threshold.
\emph{Model 4:} Here, the performance gap between DRL and both classical
strategies is maintained.

We show that in the most complex scenario, Model 4, CMort is faced with
a tradeoff---the optimal mortality rate leads a rather large fraction of
episodes ending with a population crash, whose negative reward is
counteracted with a higher economic output.\footnote{
In our mathematical formulation of the decision problem, we have assumed for simplicity that the fishing effort cost is zero and that fish price is stable over time.
This way, we equate economic output with harvested biomass.
} Conversely, more conservative, lower, mortality rates lead to lower
total reward on average. The DRL approach side-steps this trade-off by
optimizing over a more complex family of possible policies-----policies
parametrized by a neural network, as opposed to policies labeled by a
single parameter (the mortality rate).

Our findings paint a picture of how a single-species optimal management
strategy may lose performance rather dramatically when controlling a
more complex ecosystem. Here, DRL performs better from both economical
\emph{and} conservation points of view. Moreover, rather unintuitively,
CMort -- known to be suboptimal and unsustainable for single-species
models -- can turn out to even outperform the CEsc -- the single-species
optimal strategy -- for complex ecosystems. Finally, within this regime
of complex, possibly varying, ecosystems, we show that DRL consistently
finds a policy which effectively either matches the best classical
strategy, or outperforms it. We strengthen this result with a stability
investigation: we show that random perturbations in the model's
parameter values used do not significantly vary the conclusion that DRL
outperforms CEsc.

\hypertarget{mathematical-models-of-fisheries-considered}{%
\section{Mathematical models of fisheries
considered}\label{mathematical-models-of-fisheries-considered}}

Here we mathematically introduce the four fishery models for which we
compare different management strategies. In general, the class of models
that appear in this context are \emph{stochastic, first order, finite
difference equations}. For \(n\) species, these models have the general
form \begin{align}
  \label{eq:general model}
  \begin{split}
    \Delta X_t &= f_X(N_t) + \eta_{X,t} - M_X X_t\\
    \Delta Y_t &= f_Y(N_t) + \eta_{Y,t} - M_Y Y_t\\
    \Delta Z_t &= f_Z(N_t) + \eta_{Z,t} - M_Z Z_t\\
    &\dots
  \end{split}
\end{align} where
\(N_t = (X_t,\ Y_t,\ Z_t,\ \dots) \in \mathbb{R}^{n}_+\) is a vector of
populations, \(\Delta X_t := X_{t+1} - X_t\),
\(f_i:\mathbb{R}^{n}\to\mathbb{R}\) are arbitrary functions, and where
\(\eta_{i,t}\) are Gaussian random variables. Here,
\(M_i=M_i(N_t)\in\mathbb{R}_+\) is a state-dependent fish mortality
arising from harvesting the \(i\)-th species (sometimes this is referred
to as \emph{fishing effort}).

The term \(M_X X_t\) is the total \(X\) harvest at time \(t\). This
formulation of stock recruitment as a discrete finite difference process
is common among fisheries, as opposed to continuous time formulations
which involve instantaneous growth rates. This growth rate simplifies
e.g.~the possibly seasonal nature of reproduction (which would need to
be accounted for in a realistic continuous-time model) by simply
considering the total recruitment experienced by the population over a
full year.

The fishing efforts are the \emph{control variables} of our
problem---these may be set by the manager at each time-step to specified
values. We make two further simplifying assumptions on the control
problem: \emph{1. Full observation:} the manager is able to accurately
measure \(N_t\) and use that measurement to inform their decision.
\emph{2. Perfect execution:} the action chosen by the manager is
implemented perfectly (i.e., there is no noise affecting the actual
value of the fishing efforts).

Model 1 is a single-species classical model of ecological tipping
points. Models 2-4 are all three-species models with similar dynamics.
Following this logic, the first subsection will be dedicated to the
single-species model and the second will focus on the three-species
models.

\hypertarget{the-single-species-model}{%
\subsection{The single species model}\label{the-single-species-model}}

Optimal control policies for fisheries are frequently based on
1-dimensional models, \(n=1\), as described in \(Text\) \(Box\) \(1\).
The most familiar model of \(f(X)\) is that of \emph{logistic growth},
for which \begin{align}
  \label{eq:logistic}
  f(X_t) = r X_t\big(1 - X_t / K \big) =: L(X_t;\ r, K).
\end{align}

Real world ecological systems are obviously far more complicated than
this simple model suggests. One particularly important aspect that has
garnered much attention is the potential for the kind of highly
non-linear functions that can support dynamics such as alternative
stable states and hysteresis. A seminal example of such dynamics was
introduced in (May 1977), using a one-dimensional model of a prey
(resource) species under the pressure of a (fixed) predator. In the
notation of eq. \eqref{eq:general model}, \begin{align}
  \label{eq:may}
  f_X(X_t)
  = L(X_t;\ r, K) - \frac{\beta H X_t^2}{c^2 + X_t^2}.
\end{align} In the following, we will denote \begin{align*}
  F(X_t,H;\ \beta,c) := \frac{\beta H X_t^2}{c^2 + X_t^2}.
\end{align*}

The model has six parameters: the growth rate \(r\) and carrying
capacity \(K\) for \(X\), a constant population \(H\) of a species which
preys on \(X\), the maximal predation rate \(\beta\), the predation rate
half-maximum biomass \(c\), and the variance \(\sigma_X^2\) of the
stochastic term \(\eta_{X,t}\). (Here and in the following we will
center all random variables at zero.)

\begin{figure}
\includegraphics[width=6.5in]{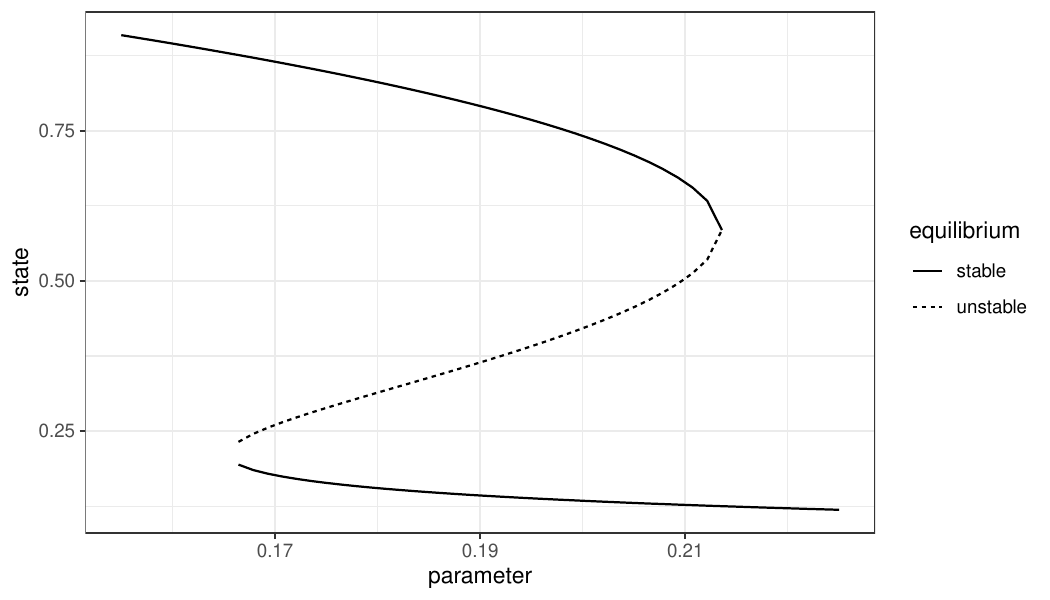} \caption{The fixed point diagram for the unharvested dynamics of Model 1 as a function of varying the parameter $\beta H$, assuming zero noise. Stable fixed points (also known as attractors) are plotted using a solid line, while the unstable fixed point is shown as a dotted line.}\label{fig:may}
\end{figure}

Eq. \eqref{eq:may} is an interesting study case of a \emph{tipping
point} (saddle-node bifurcation) (see Fig. \ref{fig:may}). Holding the
value of \(\beta\) fixed, for intermediate values of \(H\) there exist
two stable fixed points for the state \(X_t\) of the system, these two
attractors separated by an unstable fixed point. At a certain threshold
value of \(H\), however, the top stable fixed point collides with the
unstable fixed point and both are annihilated. For this value of \(H\),
and for higher values, only the lower fixed point remains. This also
creates the phenomenon of \emph{hysteresis}, where returning \(H\) to
its original value is not sufficient to restore \(X_t\) to the original
stable state.

This structure implies two things. First, that a drift in \(H\) could
lead to catastrophic consequences, with the population \(X_t\)
plummeting to the lower fixed stable point. Second, that if the
evolution of \(X_t\) is \emph{stochastic}, then, even at values of \(H\)
below the threshold point, the system runs a sizeable danger of tipping
over towards the lower stable point.

\hypertarget{the-three-species-models}{%
\subsection{The three species models}\label{the-three-species-models}}

Models 2-4 are three-species models and they are all closely
related---in fact, their natural dynamics (i.e.~dynamics under zero
harvest) is essentially given by the same equations: \begin{align}
\label{eq:3d model}
\begin{split}
  f_X(N_t) &= L(X_t;\ r_X, K_X) - F(X_t, Z_t;\ \beta, c) - c_{XY}X_tY_t, \\
  f_Y(N_t) &= L(Y_t;\ r_Y, K_Y) - D F(Y_t, Z_t;\ \beta, c) - c_{XY}X_tY_t,\\
  f_Z(N_t) &= \left(b(X_t + DY_t\right) - d_Z)Z_t.
\end{split}
\end{align} The three species modeled are \(X\), \(Y\) and \(Z\).
Species \(Z\) preys on both \(X\) and \(Y\), while the latter two
compete for resources. There are thirteen parameters in this model: The
growth rate and carrying capacity, \(r_X\), \(K_X\), \(r_Y\) and
\(K_Y\), of \(X\) and \(Y\). A parameter \(c_{XY}\) mediating a
Lotka-Volterra competition between \(X\) and \(Y\). A maximum predation
rate \(\beta\) and a predation rate half-maximum biomass \(c\)
specifying how \(Z\) forages on \(X\) and \(Y\). A parameter \(D\)
regulating a relative preference of \(Z\) to prey on \(Y\). A death rate
\(d_Z\) and a parameter \(b\) scaling the birth rate of \(Z\). Finally,
the noise variances \(\sigma_X\), \(\sigma_Y\) and \(\sigma_Z\).

The three models will branch off of eq. \eqref{eq:3d model} in the
following way. \emph{Model 2:} here, only \(X\) is harvested, that is,
in the notation of eq. \eqref{eq:general model}, we fix \(M_Y=M_Z=0\)
and leave \(M_X\) as a control variable. All parameters here are
constant. \emph{Model 3:} as Model 1, but with \(X\) and \(Y\) being
harvested. In other words, we set \(M_Z=0\) and leave \(M_X\) and
\(M_Y\) as control variables. \emph{Model 4:} here \(X\) and \(Y\) are
harvested, but now we include a non-stationary parameter:
\begin{equation}
  \label{eq:rx(t)}
  r_X = r_X(t) = 
  \begin{cases}
    1-t/200, \quad & t \leq 100,\\
    1/2, \quad & t > 100.
  \end{cases}
\end{equation} All other parameters are constant. Eq. \eqref{eq:rx(t)}
is intended to reflect in a simple manner a possible effect of climate
change: where the reproductive rate of \(X\) is reduced linearly over
time until it stabilizes.

\hypertarget{sec:RL}{%
\section{Reinforcement learning}\label{sec:RL}}

Reinforcement learning (RL) is a way of approaching \emph{sequential
decision problems} through machine learning. All applications of RL can
be conceptually separated into two parts: an \emph{agent} and an
\emph{environment} which the agent interacts with. That is, the agent
performs actions within the environment.

After the agent takes an action, the environment will transition to a
new state and return a numerical \emph{reward} to the agent. (See Fig. 1
in (Lapeyrolerie et al. 2022) for a conceptual description of
reinforcement learning algorithms.) The rewards encode the agent's goal.
The main task of any RL algorithm is then to maximize the cumulative
reward received. This objective is achieved by aggregating experience in
what is called the \emph{training} period and learning from such
experience.

The \emph{environment} is commonly a computer simulation. It is
important to note here the role that real time-series data of stock
sizes can play in this process. This data is not used directly to train
the RL agent, but rather to estimate the model defining the environment.
This environment is subsequently used to train the agent. In this paper,
we focus on the second step---we take the estimated model of reality as
a given, and train an RL agent on it.\footnote{
In this sense, it is important to note that the classical management strategies we compare against have a similar flow of information.
Namely, data is used to estimate a dynamical model, and this model is used to generate a policy function.
The difference to our approach is located in the process of *how* the model is used to optimize a policy.
Because of this difference, RL-based approaches can produce good heuristic solutions for complex problems.
}

Specifically, we consider four environments corresponding to each of the
models considered (Table \ref{tab:models}). At each time step, the agent
observes the state \(S\) and enacts some harvest---reducing \(X_t\) to
\(X_t - M_X(N_t)\cdot X_t\), and, for Models 3 and 4, also reducing
\(Y_t\) to \(Y_t - M_Y(N_t)\cdot Y_t\). Here the fish
mortality-rates-from-harvest (i.e.~\(M_X=M_X(N_t)\) and
\(M_Y=M_Y(N_t)\)), are the agent's action at time \(t\). This secures a
reward of \(M_X(N_t)X\) for Models 1 and 2, and, similarly, a reward of
\(M_X(N_t)X + M_Y(N_t)Y\) for Models 3 and 4. After this harvest portion
of the time step, the environment evolves naturally according to eqs.
\eqref{eq:may} and \eqref{eq:3d model} (Sec. 2).

As mentioned previously, discretising time allows a simplification of
the possibly seasonal mating behavioral patterns of the species
involved. This approximation is commonly used in fisheries for species
with annual reproductive cycles (see e.g. (Mangel 2006), Chap. 6).
Moreover, the separation of each time-step into a harvest period and a
natural growth period assumes that harvest has little disruption on the
reproductive process. A detailed model which includes such a disruption
is outside of the scope of this work.

\hypertarget{mathematical-framework-for-rl}{%
\subsection{Mathematical framework for
RL}\label{mathematical-framework-for-rl}}

The RL environment can be formally described as a discrete time
\emph{partially observable Markov decision process (POMDP)}. This
formalization is rather flexible and allows one, e.g., to account for
situations where the agent may not fully observe the environment state,
or where the only observations available to the agent are certain
functions of the underlying state. For the sake of clarity, we will only
present here the subclass of POMDPs which are relevant to our work:
\emph{fully observable MDPs} (henceforth MDPs for short). An MDP may be
defined by the following data:

\begin{itemize}
\item
  \(\mathcal{S}\): \emph{state space}, the set of states of the
  environment,
\item
  \(\mathcal{A}\): \emph{action space}, the set of actions which the
  agent may choose from,
\item
  \(T(N_{t+1}|N_t, a_t, t)\): \emph{transition operator}, a conditional
  distribution which describes the dynamics of the system (where
  \(N_i\in\mathcal{S}\) are states of the environment),\footnote{
  Transition operators are commonly discussed without having a direct time-dependence for simplicity, but the inclusion of $t$ as an argument to $T$ does not alter the structure of the learning problem appreciably.
  }
\item
  \(r(N_t, a_t, t)\): \emph{reward function}, the reward obtained after
  performing action \(a_t\in\mathcal{A}\) in state \(N_t\),
\item
  \(d(N_0)\): \emph{initial state distribution}, the initial state of
  the environment is sampled from this distribution,
\item
  \(\gamma\in[0,1]\): \emph{discount factor}.
\end{itemize}

At a time \(t\), the MDP agent observes the full state \(s_t\) of the
environment and chooses an action based on this observation according to
a \emph{policy function} \(\pi(a_t | N_t)\). In return, it receives a
discounted reward \(\gamma^t r(a_t, N_t)\). The discount factor helps
regularize the agent, helping the optimization algorithm find solutions
which pay off within a timescale of \(t \sim \log(\gamma^{-1})^{-1}\).

With any fixed policy function, the agent will traverse a path
\(\tau=(N_0,\ a_0,\ N_1,\ a_1\ \dots,\ N_{t_{\text{fin.}}})\) sampled
randomly from the distribution \begin{align*}
  p_\pi(\tau) = 
  d(N_0) \prod_{t=0}^{ t_{\text{fin.}}-1 }
  \pi( a_t | N_t ) T( N_{t+1} | N_t, a_t, t ).
\end{align*} Reinforcement learning seeks to optimize \(\pi\) such that
the expected rewards are maximal, \[  
  \pi^* = \mathrm{argmax}\ \mathbb{E}_{\tau\sim p_\pi}[R(\tau)],
\] where, \[
  R(\tau) = \sum_{t=0}^{ t_{\text{fin.}}-1 } \gamma^t r(a_t, N_t, t),
\] is the cumulative reward of path \(\tau\). The function
\(J(\pi):=\mathbb{E}_{\tau\sim p_\pi}[R(\tau)]\) is called the
\emph{expected return}.

\hypertarget{deep-reinforcement-learning}{%
\subsection{Deep Reinforcement
Learning}\label{deep-reinforcement-learning}}

The optimal policy function \(\pi\) often lives in a high or even
infinite-dimensional space. This makes it unfeasible to directly
optimize \(\pi\). In practice, an alternative approach is used: \(\pi\)
is optimized over a much lower-dimensional parameterized family of
functions.\footnote{
Policies are, in general, functions from state space to policy space.
In our paper, these are $\pi:[0,1]^{\times 3}\to \mathbb{R}_+$ for the single fishery case, and $\pi:[0,1]^{\times 3}\to \mathbb{R}_+^2$ for two fisheries.
The space of all such functions is highly singular, spanning a \emph{non-separable Hilbert space}.
Even restricting ourselves to continuous policy functions, we end up with a set of policies which span the infinite dimensional space $L^2([0,1]^{\times3})$.
One way to avoid optimizing over an infinite dimensional ambient space is to discretize state space into a set of bins.
This approach runs into tractability problems: First, the dimension of policy space scales exponentially with the number of species.
Second, even for a fixed number of species (e.g., 3), the dimension optimized over can be prohibitively large---for example if one uses 1000 bins for each population in a three-species model, the overall number of parameters being optimized over is $10^9$.
Neural networks with much smaller number of parameters, on the other hand, can be quite expressive and sufficient to find a rather good (if not optimal) policy function.
} Deep reinforcement learning uses this strategy, focusing on function
families parameterized by neural networks. (See Fig. 1 and App. A in
(Lapeyrolerie et al. 2022) for a conceptual introduction to the use of
reinforcement learning in the context of conservation decision making.)

We will focus on deep reinforcement learning throughout this paper.
Within the DRL literature there is a wealth of algorithms from which to
choose to optimize \(\pi\), each with its pros and cons. Most of these
are based on gradient ascent by using the technique of
\emph{back-propagation} to efficiently compute the gradient. Here we
have used only one such algorithm (\emph{proximal policy optimization
(PPO)}) to draw a clear comparison between the RL-based and the
classical fishery management approaches. In practice, further
improvements can be expected by a careful selection of the optimization
algorithm. (See, e.g., (François-Lavet et al. 2018) for an overview of
different optimization schemes used in DRL.)

\hypertarget{model-free-reinforcement-learning}{%
\subsection{Model-free reinforcement
learning}\label{model-free-reinforcement-learning}}

Within control theory, the classical setup is one where we use as much
information from the model as possible in order to derive an optimal
solution. Here, one may find a vast literature on model-based methods to
attain optimal, or near-optimal, control (see, e.g., (Zhang, Li, and
Liao 2019; Sethi and Sethi 2019; Anderson and Moore 2007)).

The classical sustainable fishery management approaches summarized in
Text Box 1, for instance, are model-based controls. As we saw there,
these controls may run into trouble in the case where there are
inaccuracies in the model parameter estimates.

There are many situations, however, in which the exact model of the
system is not known or not tractable. This is a standard situation in
ecology: mathematical models capture the most prominent aspects of the
ecosystem's dynamics, while ignoring or summarizing most of its
complexity. In this case, it is clear, model-based controls run a grave
danger of mismanaging the system.

Reinforcement learning, on the other hand, can provide a model-free
approach to control theory. While a model is often used to generate
training data, this model is not directly used by model-free RL
algorithms. This provides more flexibility to use RL in instances where
the model of the system is not accurately known. In fact, it has been
shown that model-free RL outperforms model-based alternatives in such
instances (Janner et al. 2019). (For recent surveys of model-based
reinforcement learning, which we do not focus on here, see (Moerland et
al. 2023; Polydoros and Nalpantidis 2017).)

This context provides a motivation for this paper. Indeed, models for
ecosystem dynamics are only ever approximate and incomplete descriptions
of reality. This way, it is plausible that model-free RL controls could
outperform currently used model-based controls in ecological management
problems.

Model-free DRL provides, moreover, a framework within which agents can
be trained to be generally competent over a \emph{variety} of different
models. This could more faithfully capture the ubiquitous uncertainty
around ecosystem models. The aforementioned framework---known as
\emph{curriculum learning}---is considerably more intensive on
computational resources than the ``vanilla'' DRL methods we have used in
this paper.\footnote{
All our agents were trained in a local server with two commercial GPUs. 
The training time was between 30 minutes and one hour in each case.
} Due to the increased computational requirements of this framework, we
have left the exploration in this direction for future work.

\hypertarget{methods}{%
\section{Methods}\label{methods}}

\textbf{The Environment.} We considered the problem of managing four
increasingly complex models (see Table \ref{tab:models}). To recap,
these four environments are a single-species growth model \eqref{eq:may}
for a harvested species; a three-species model \eqref{eq:3d model} with
a single harvested species; the same three-species model but with
\emph{two} harvested species; and, finally, a three-species model with a
time-varying parameter and two harvested species.

The policies explored were functions from states \(N_t\) to either a
single number \(M_X(N_t) = \pi(N_t)\) (for Models 1 and 2), or a pair of
numbers \((M_X(N_t), M_Y(N_t)) = \pi(N_t)\) (for Models 3 and 4).

Our goal was to evaluate the performance of different policy strategies
over a specified window of time. We chose this window to be 200
time-steps, where each discrete time-step represents the dynamical
evolution of the system over a year. Each time-step was composed of two
parts: First, a \emph{harvest period} where the harvest is collected
from the system (e.g.~the population \(X\) is reduced to
\(X_t\mapsto X_t - M_X(N_t) X_t\)). Second, a \emph{recruitment and
interaction period}, where the system's state evolves according to its
natural dynamics.

Training and evaluating management strategies were performed by
simulating \emph{episodes.} An episode begins at a fixed initial state
and the system is controlled with a management policy until \(t=200\),
or until a ``near-extinction event'' occurs---that is, until any of the
populations go below a given threshold, \begin{align}
  \label{eq:thresh}
  X_t \leq X_{\text{thresh.}},\quad
  Y_t \leq Y_{\text{thresh.}},\quad
  \text{or,}\quad
  Z \leq Z_{\text{thresh.}}.
\end{align} In our setting we have chosen
\(X_{\text{thresh.}}=Y_{\text{thresh.}}=Z_{\text{thresh.}}=0.05\) as a
rule of thumb---given that under natural (unharvested) dynamics the
populations range within values of 0.5 to 1, this would represent on the
order of a 90-95\% population decrease from their ``natural'' level.

The reward function defining our policy optimization problem had two
components. The first was economical: the total biomass harvested over
an episode. The second seeked to reflect conservation goals: if a
near-extinction event occurred at time \(t\), the episode was ended
early and a negative reward of \(-100/t\) was awarded. This reward
function balanced the extractive motivation of the fishery with
conservation goals which went beyond the scope of long-term sustainable
harvests commonly used in fishery management.

\textbf{Training a DRL agent.} We trained a DRL agent parametrized by a
neural network with two hidden, 64-neuron, layers, on a local server
with 2 commercial GPUs. We used the Ray
framework\footnote{\url{https://docs.ray.io/}} for training,
specifically we used the Ray PPOConfig class to build the policy
optimization algorithm using the default values for all
hyper-parameters. In particular, no hyperparameter tuning was performed.
The agent was trained for 300 iterations of the PPO optimization step
for the three-species cases. The total training time was on the order of
30 minutes to 1 hour. For the single-species model, the training
iterations were scaled down to 100 and the training time was around 10
minutes.

The state space used was normalized case-by-case as follows: \emph{Model
1:} a line segment \([0,1]\), \emph{Models 2-4:} a cube
\([0,1]^{\times 3}\). We used simulated data to derive a bound on the
population sizes typically observed, and thus be able to normalize
states to a finite volume.

Policies obtained from the PPO algorithm can be ``noisy'' as their
optimization algorithm is randomized (see, e.g.~App. B for a
visualization of the PPO policy obtained for Model 4). We smoothed this
policy out using a Gaussian process regressor interpolation. Details for
this interpolation process can be found in App. C.

\textbf{Tuning the CMort strategy.} In order to estimate the optimum
mortality rate, we optimized over a grid of possible mortality rates.
Namely, for Models 1 and 2, grid of 101 mortality rates was laid in the
interval \([0, 0.5]\); for Models 3 and 4, a \(51\times51\) grid was set
in the square \([0,0.5]^2\). The latter had a slightly coarser grid due
to the high memory cost of using the denser grid. Since the approach for
tuning was completely analogous for all models evaluated, here we
discuss only Model 4. For each one of these choices of mortality rates,
say \((M_X, M_Y)\), we simulated 100 episodes based on
\eqref{eq:3d model}: At each time step the state \((X_t, Y_t, Z_t)\) was
observed, a harvest of \(M_X X_t+M_YY_t\) was collected and then the
system evolved according to its natural dynamics \eqref{eq:3d model}.
The optimal mortality rate was the value \((M_X^*,\ M_Y^*)\) for which
the mean episode reward was maximal.

\begin{figure}
  \centering
  \includegraphics[scale=0.5]{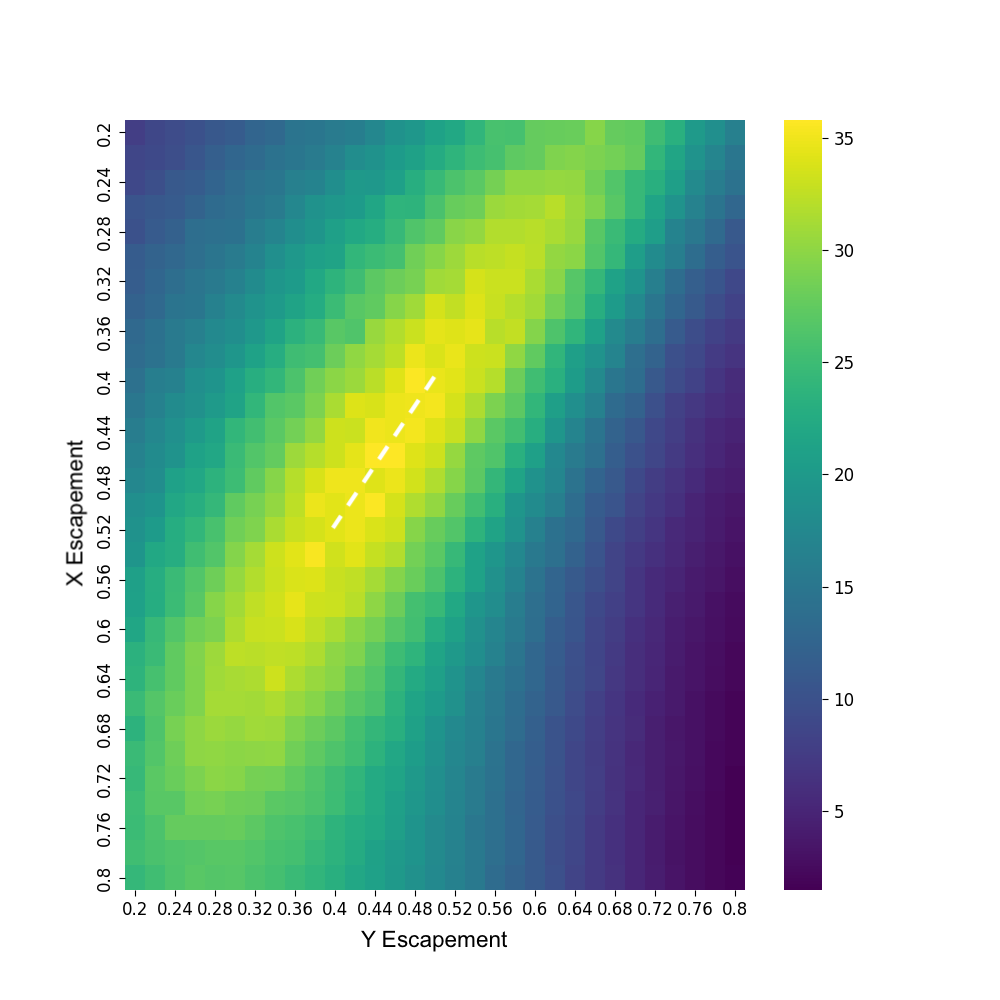}
  \caption{Visualization of the cosntant escapement strategy tuning procedure for Model 4. There was a certain multiplicity in this tuning strategy: a ‘‘ridge of optimality’’ where policies had essentially equivalent behavior. Throughout our investigation, we tuned constant escapment in several occasions and, on each occassion, a different optimal policy along the ridge was found. The results for different policies along the ridge were in practice equivalent, with no discernible difference in performance. We highlighted the ridge with a white dotted line.}
  \label{fig:tuning}
\end{figure}

\textbf{Tuning the CEsc strategy.} This tuning procedure was analogous
to that of the CMort strategy just summarized. Namely: A grid of 101
escapement values was laid out on the interval \([0,1]\) for Models 1
and 2; and a \(51\times 51\) grid on \([0,1]^2\) was laid out for Models
3 and 4. Each grid point represented a CEsc policy. We used each of
these policies to manage 100 replicate episodes. The optimal policy was
the policy with the highest average reward obtained. A visualization of
the tuning outcome for Model 4 is shown in Fig. \ref{fig:tuning}.

\textbf{Parameter values used.} The single-species model's (eq.
\eqref{eq:may}) dynamic parameters were chosen as \begin{align}
\label{eq:1d param vals}
r = K = 1,\quad \beta = 0.25,\quad c = 0.1.
\end{align} Here, the values of \(\beta\) and \(c\) were chosen as to
make the system be roughly close to its tipping point.

For Models 2 and 3, their dynamic parameters (in eq.
\eqref{eq:3d model}) were chosen as follows \begin{align}
\label{eq:param vals}
\begin{split}
  r_X = K_X = r_Y = K_Y = 1, \quad \beta = 0.3, \quad c = 0.3\\
  c_{XY} = 0.1,\quad b = 0.1,\quad D = 1.1, \quad d_Z = 0.1.
\end{split}
\end{align} Moreover, the variances for the stochastic terms were
chosenas \[
  \sigma^2(\eta_{X,t}) = \sigma^2(\eta_{Y,t}) = \sigma^2(\eta_{Z,t}) = 0.05.
\] For Model 4, we used \(r_X(t)\) given as in eq. \eqref{eq:rx(t)} and
all other parameters were given as in eq. \eqref{eq:param vals}.

The values of \(c\) and \(\beta\) in the three-species model were
slightly different than their values in the one-species model. These
values were chosen heuristically: We observed that choosing the lower
value of \(c=0.1\) in this case would lead to quick near-extinction
events even without a harvest. Moreover, \(\beta\) was slightly
increased simply to put more predation pressure on the \(X\) and \(Y\)
populations and make them slightly more fragile.

\textbf{Stability analysis.} To ensure that our results do not strongly
depend on our parameter choices, we performed a stability analysis.
Here, we perturbed parameters randomly and measured the difference in
performance between our DRL-based methods and the CEsc strategy. We
observed that the difference in performance is maintained for even
relatively high noise strengths. We only considered the most complex
case here, Model 4.

For each value of parametric noise strength \(\sigma_{\text{param.}}\),
we executed the following procedure: We sampled 100 choices of perturbed
parameter values, where the perturbation was as follows---each parameter
\(P\) was perturbed to \((1+g_P)P\) where \(g_P\) was a Gaussian random
variable with variance \(\sigma_{\text{param.}}^2\). For each of these
sample parameter sets, we tuned CEsc and trained the DRL agent. We
measured the average reward difference between these two strategies for
each of sample (this was done using 100 replicate evaluation episodes).
Finally, we took the mean of this average difference over the 100
perturbed parameter samples.

The parametric noise strength values used were
\([0.04, 0.08, \dots, 0.2]\).

\hypertarget{results}{%
\section{Results}\label{results}}

We evaluated each of the four management strategies considered on Models
1-4. To recap, the management strategies were CEsc, CMort, PPO and a
Gaussian process interpolation of PPO (``PPO+G''). Furthermore, we
characterized the trade-off between economic output and sustainability
faced by CMort policies. This was done by evaluating CMort policies with
a fraction of the optimal mortality rate (specifically, 80\%, 90\% and
95\%). All evaluations were based on 100 replicate episodes.

We will visualize the results concerning Model 4 in this section,
leaving the other models for App. A. This is the most complex scenario
considered and where our results show the most compelling advantage of
DRL methods with respect to classical strategies.

\begin{figure}
\includegraphics[width=6in]{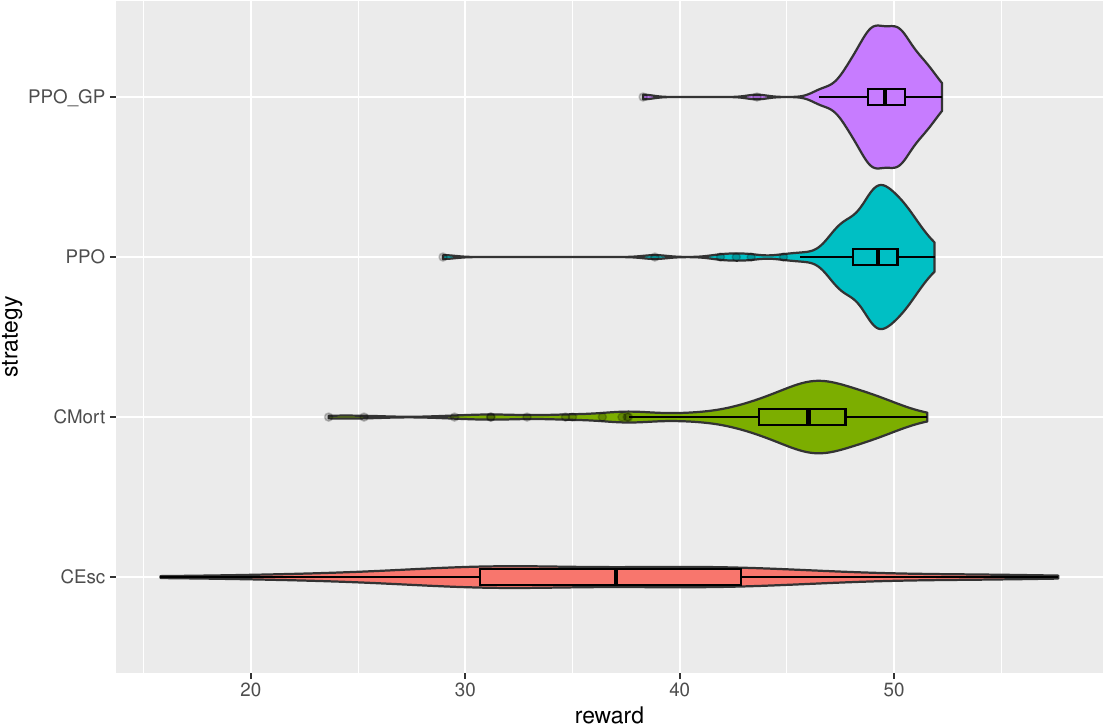} \caption{Reward distributions for the four strategies considered. These are based on 100 evaluation episodes. We denote CEsc for constant escapement, CMort for constant mortality, PPO for the output policy of the PPO optimization algorithm, and PPO GP for the Gaussian process interpolation of the PPO policy.}\label{fig:main}
\end{figure}

Our main result is summarized in Fig. \ref{fig:main} which displays the
total reward distributions for the policy obtained through each
strategy. Here we see that CEsc has a long-tailed distribution of
rewards, and its average reward is much lower than other management
strategies. CMort has a shorter-tailed distribution and a much higher
average reward. Finally, both DRL-based strategies have a more
concentrated reward distribution and a higher average reward than CMort.

\begin{figure}
\includegraphics[width=6in]{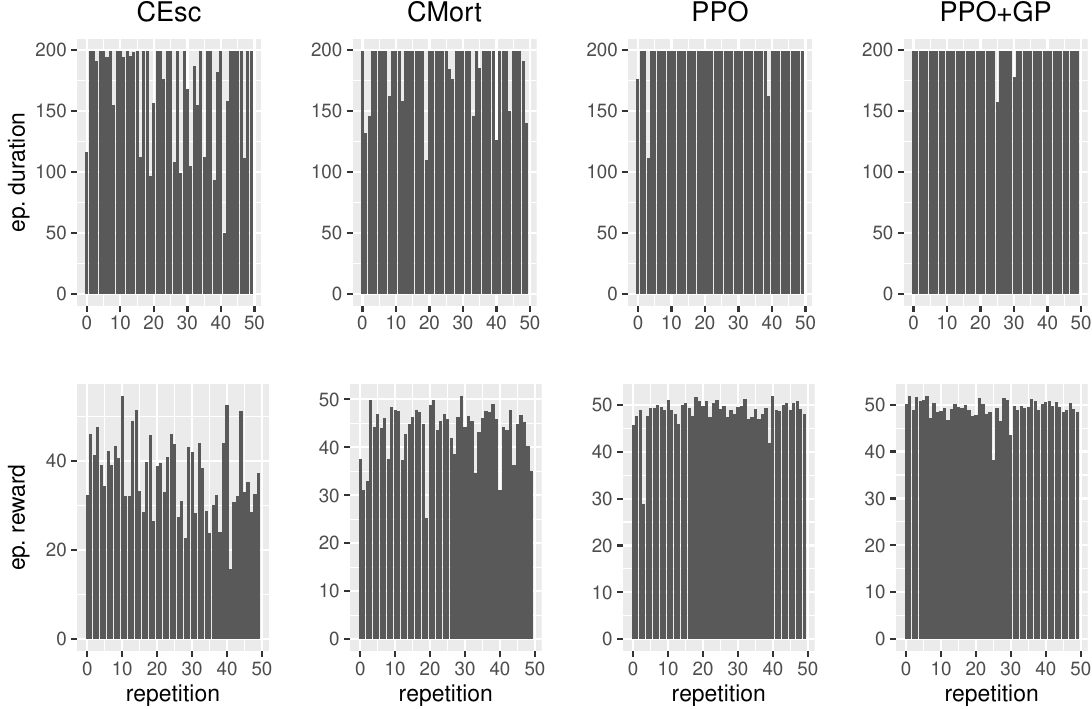} \caption{Hisotgrams of episode lengths and rewards for the four different management strategies considered. Only the first 50 evaluation episodes (from a total of 100) were included, for ease of visualization. From left to right, the four management strategies compared are CEsc, CMort, PPO, and PPO+GP.}\label{fig:ephistograms}
\end{figure}

To assess what is the culprit for the classical strategies' low
performance with respect to the DRL-based strategies, we plot the
duration of each of the 100 evaluation episodes of each strategy in Fig.
\ref{fig:ephistograms}. We see that early episode ends are prevalent in
classical strategies and rare for DRL-based strategies. Early episode
ends tend to happen at lower \(t\) values for CEsc than CMort. Thus,
distribution of episode durations seems to be widest for CEsc, followed
by CMort, DRL and DRL+GP.

\begin{figure}
\includegraphics[width=6in]{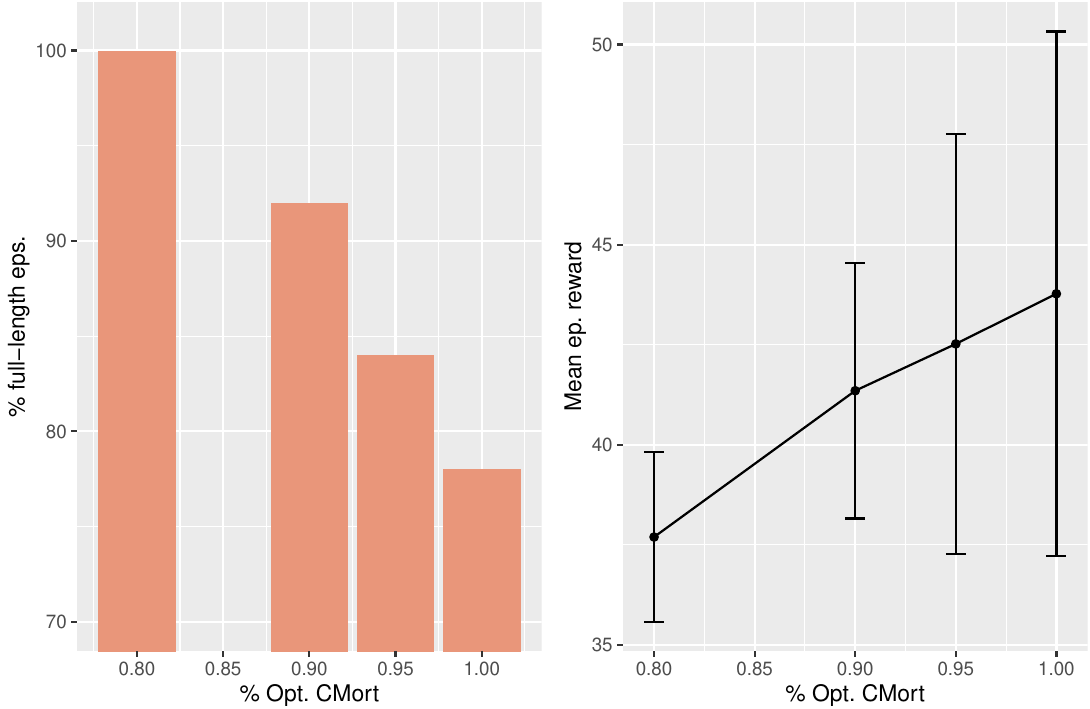} \caption{Trade-off between reward and probability of a near-extinction event for CMort policies. We evaluated policies at the full optimal constant escapement value, and also at 0.8, 0.9, 0.95 of the latter. Each evaluation is based on 100 episodes. On the left we plot the percentage of episodes which last their maximum time window, i.e. that do not see a near-extinction event. On the right, we plot the mean episdoe reward and standard deviation for each policy.}\label{fig:fracmsy}
\end{figure}

We examine the trade-off between profit and sustainability faced by the
CMort strategy in Fig. \ref{fig:fracmsy}. Two quantities are plotted:
the fraction of evaluation episodes with maximal length (i.e.~episodes
with no near-extinction events), and the average reward. On the x-axis
we have several sub-optimal mortality rates that err on the conservative
side: e.g.~the policy labeled ``80\% Opt. CMort'' has the form
\begin{align*}
  \pi: (X_t, Y_t, Z_t)\mapsto (0.8 M_X^*,\ 0.8 M_Y^*),
\end{align*} where \((M_X^*,\ M_Y^*)\) is the optimal CMort strategy. We
see that sufficiently conservative policies attain high sustainability,
but only at a high price in terms of profit.\footnote{
As noted before, here we equate economic profit with biomass caught.
This is done as an approximation to convey the conceptual message more clearly, and we do not expect our results to significantly change if, e.g., ‘‘effort cost’’ is included in the reward function.
When we refer to ‘‘large differences’’ in profit, or ‘‘paying dearly,’’ we mean that the ratio between average rewards is considerable---e.g. a 15\% loss in profit.
} We expect a similar and, possibly, more pronounced effect for the CEsc
strategy but do not analyze this case here.

One problem that is often encountered when using machine learning
methods is the interpretability of the output of these methods. For our
PPO strategy, the output is a policy function parametrized by a neural
network \(\theta\): \begin{align*}
  \pi_\theta: N_t \mapsto (M_X(N_t), M_Y(N_t)),
\end{align*} where \(N_t=(X_t,Y_t,Z_t)\) is the state of the system at
time \(t\), and where \(M_X\) and \(M_Y\) are the mortalities due to
harvest during that time-step. While the values of the neural network
parameters are hard to interpret, the actual shape of the policy
function is much more understandable.

Here we visualize the PPO+GP policy function and provide an
interpretation for it, as this function is smoother and less noisy than
the PPO policy function. The PPO policy function is visualized similarly
in App. B.

\begin{figure}
  \centering
  \includegraphics[scale=0.25]{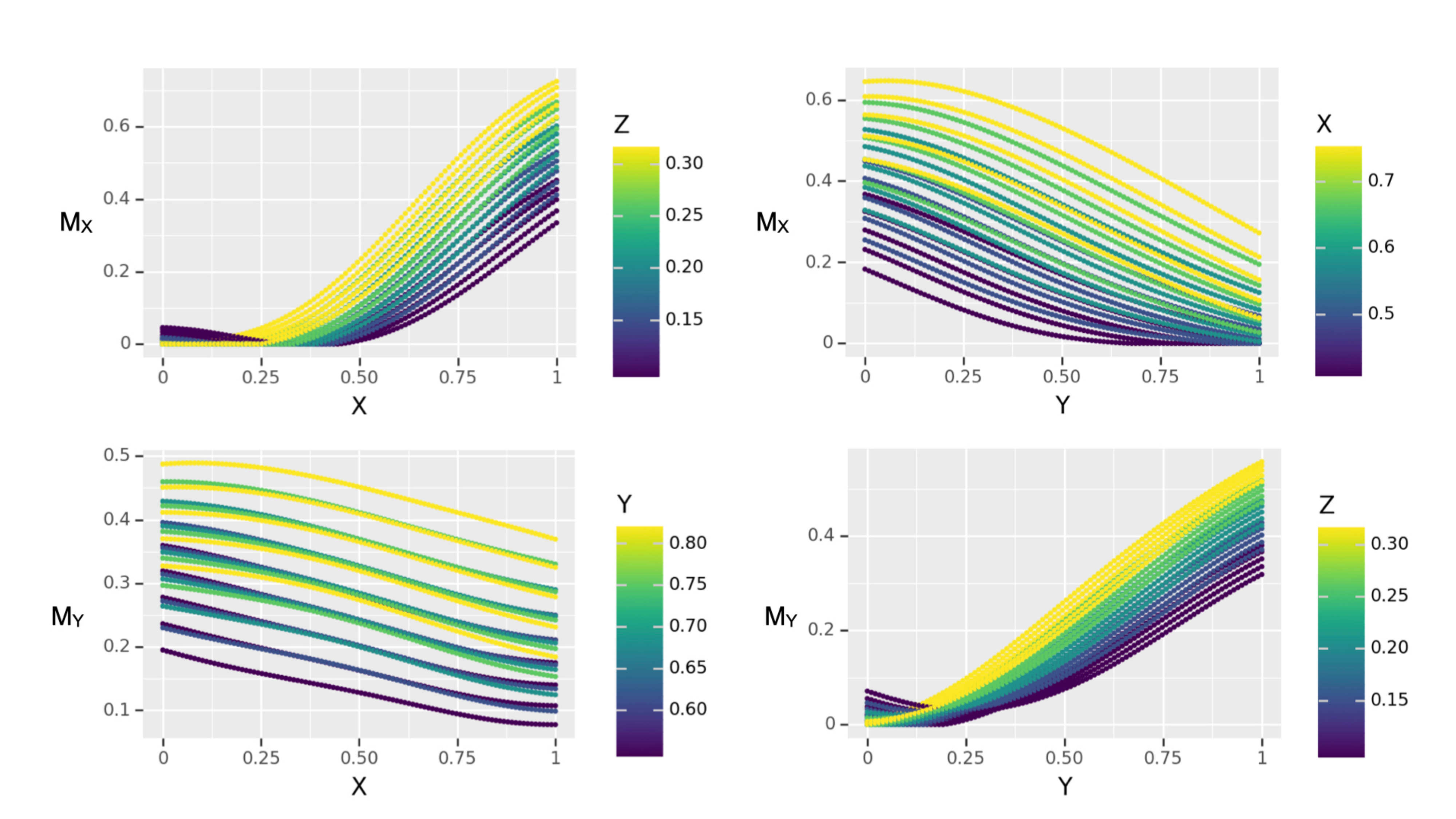}
  \caption{Plots of the PPO+GP policy $\pi_{\text{PPO+GP}}$ along several relevant axes. Here $M_X$ and $M_Y$ are the $X$ and $Y$ components of the policy function. The values of the plots are generated in the following way: For each variable $X$, $Y$, and $Z$, the time-series of evaluation episodes are used to generate a window of typical values that variable attains when controlled by $\pi_{\text{PPO+GP}}$. Then, for each plot either $X$ or $Y$ was varied on $[0,1]$ along the $x$ axis, while the other variables (resp. $Y$ and $Z$, or $X$ and $Z$) were varied within the typical window computed before. The value of one of the latter two variables were visualized as color.}
  \label{fig:gpp-policy}
\end{figure}

Given its high dimension, it is not possible to fully display how the
policy function obtained ``looks like''---we thus project it down to
certain relevant axes. The result of this procedure is shown in Fig.
\ref{fig:gpp-policy}. In that figure, the shape of the optimal
escapement strategy is provided for comparison.

We notice that the DRL-derived policy has similarities to a CEsc policy.
Here, the key difference is that the escapement value for each of the
fished species is sensitive to variations in the other populations. This
can be seen as color gradients in the plots of \((X, M_X)\) and
\((Y, M_Y)\), where the gradient corresponds to differing values of
\(Z\). Moreover, this can be seen as an anti-correlation in the plot of
\((X, M_Y)\)---for optimal CEsc, \(M_Y\) is uncorrelated to \(X\).

\begin{figure}
  \centering
  \includegraphics[scale=0.33]{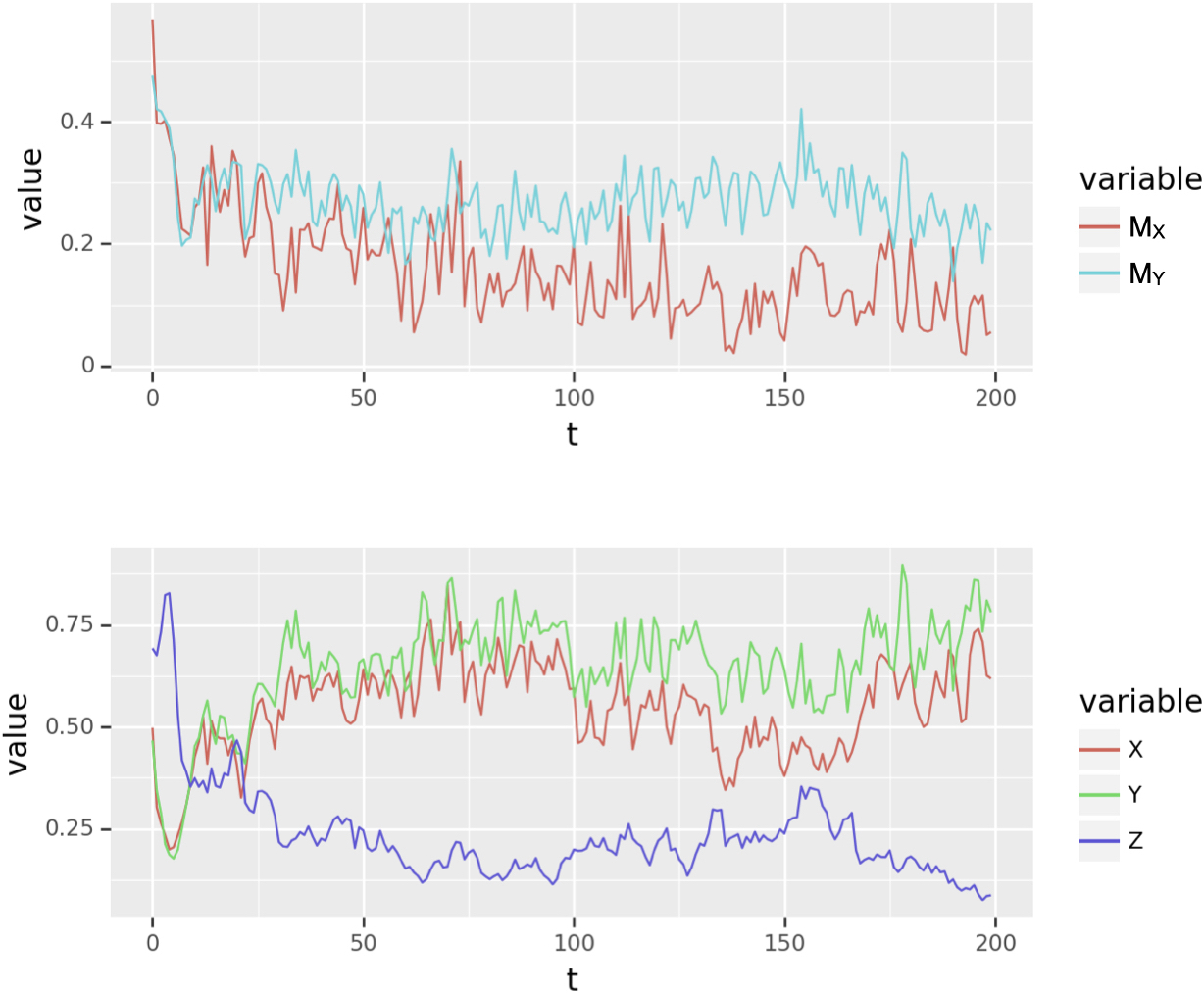}
  \caption{Time-series of an episode managed with $\pi_{\text{PPO+GP}}$. Here we plot the state of the system on the bottom panel, and the actions taken (fishing efforts on $X$ and $Y$---respectively $M_X$ and $M_Y$) on the top panel.}
  \label{fig:gpp-ep}
\end{figure}

This sensitivity of the policy to, for instance, the values of \(Z\) can
be seen in the sample time series displayed in Fig. \ref{fig:gpp-ep}.
Here, we can see that species \(Z\) becomes endangered due to harvesting
for all management strategies. The DRL-based strategy, however, is
sensitive to the values of \(Z\) and can respond accordingly by scaling
the fishing effort with the value of \(Z\). In particular, the policy
responds to the period of diminishing values of \(Z\) near the end of
the episode, by restricting fishing on \(X\) and \(Y\), thus promoting
\(Z\)'s growth. This pattern is rather common among the whole
dataset---early episode ends are largely due to near-extinctions of
\(Z\) for all management strategies.

\hypertarget{recovering-constant-escapement-for-a-single-species}{%
\subsection{Recovering constant escapement for a single
species}\label{recovering-constant-escapement-for-a-single-species}}

\begin{figure}
  \centering
  \includegraphics[scale=0.25]{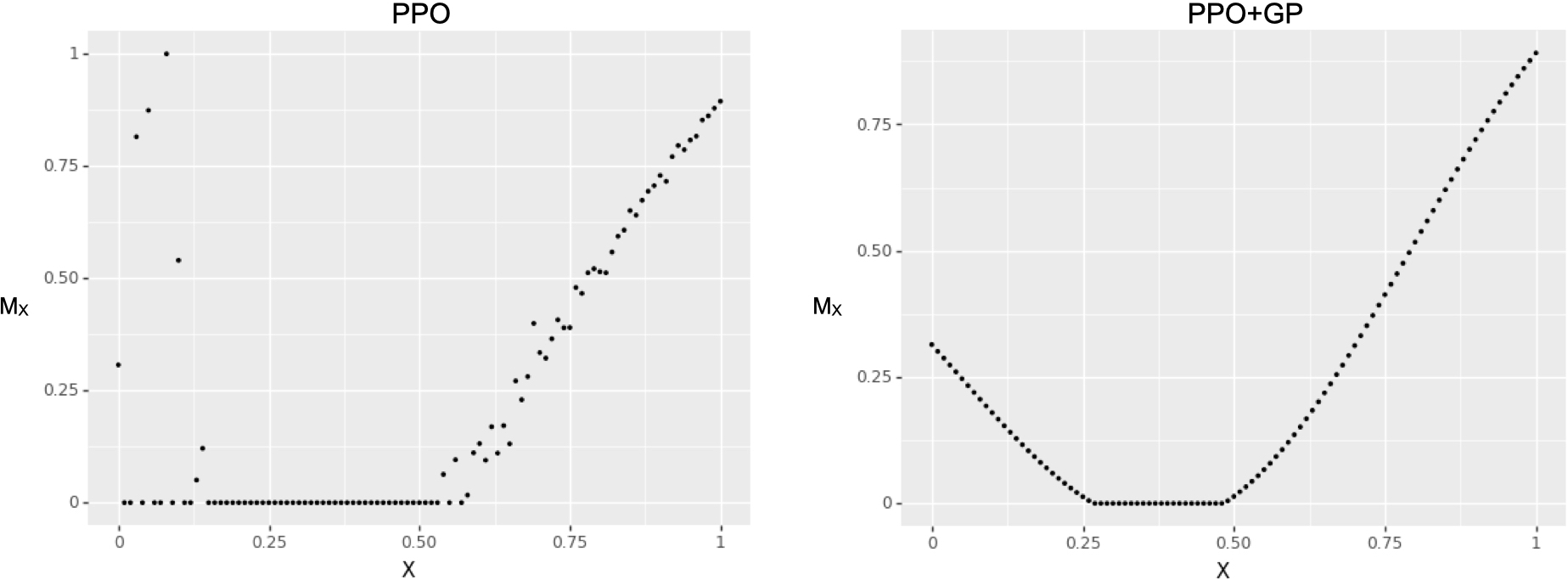}
  \caption{Left panel: the policy obtained from 100 training iterations of the PPO algorithm on the ‘‘single species, single fishery’’ model. Right panel: the Gaussian process interpolation of the left panel. We plot both as scatter data evaluated on a 101-point grid on $[0,1]$, but these policies may of course be evaluated continuously---on any possible value of $X$.}
  \label{fig:1sp-pol}
\end{figure}

While the optimal control for our single-species model \eqref{eq:may}
can not be easily proven to be CEsc (since the right-hand side of that
equation is not concave), from experience we can expect CEsc to either
be optimal or near-optimal. We give evidence for this intuition by
showing that our DRL method recovers a CEsc solution when trained. These
results are shown in Fig. \ref{fig:1sp-pol} Here we show both the output
PPO policy, and its Gaussian process interpolation. This helps build an
intuition about the relationship between our ``PPO'' and ``PPO+GP''
management strategies.

There is a presence of certain high-mortality points at low \(X\) values
in the PPO policy (which in turn generates a rising fishing mortality
for \(X\) values below a certain threshold in the PPO+GP policy). This
is likely due to experience of near-extinctions early on in the training
process---where, given an impending extinction, there is a higher reward
for intensive fishing. These ``jitters'' are likely not fully erased
through the optimization algorithm since near-extinction events become
extremely rare after only a few training iterations. This way, the agent
does not further explore that region of state space to generate new
experience. We believe the most important aspect of the CEsc policy
reproduced by PPO is the fact that there is some sufficiently-wide
window below the threshold of the policy (i.e.~below the optimal
escapement value), on which no fishing is performed. That is, there
exists some sufficiently large \(\varepsilon\) such that if
\(X_{\text{thresh.}}\) is the optimal escapement value of the system,
then \(\pi_{PPO}(X)=0\) for all
\(X\in[X_{\text{thresh.}}-\varepsilon,\ X_{\text{thresh.}}]\).

\hypertarget{stability-analysis}{%
\subsection{Stability analysis}\label{stability-analysis}}

In this section we present results intended to show that the effects
that we observe in this paper are not the result of a careful selection
of parameter values, but rather arise for a wide variety of parameter
values.

\begin{figure}
  \centering
  \includegraphics[scale=0.8]{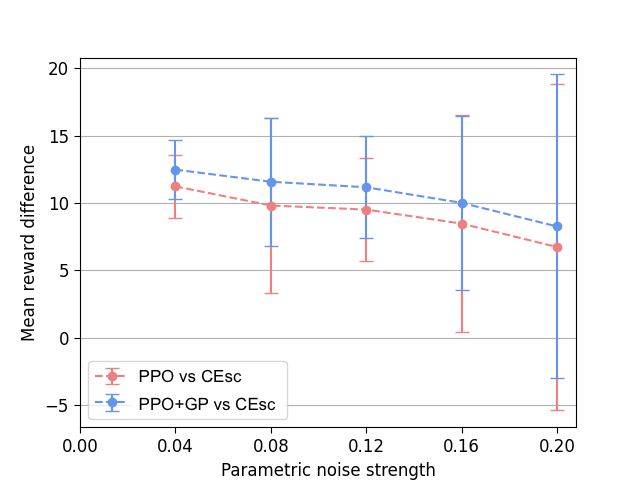}
  \caption{Mean reward difference between DRL methods (resp. our ‘‘PPO’’ and ‘‘PPO+GP’’ strategies), on the one hand, and the optimal constant escapement policy (‘‘CEsc’’) on the other. The dynamic parameters in eq. \eqref{eq:3d model} were randomly perturbed from the values given in \eqref{eq:param vals} according to the procedure detailed in Sec. 4. The noise strength of this perturbation is plotted as the x-axis. For each noise strength, 100 parameter perturbations were sampled---each one giving rise to a \emph{realization} of the model. For each such realization, we optimized a CEsc policy and trained a PPO agent. Moreover, we interpolated the PPO policy using a Gaussian process, as detailed in Sec 4. Then, for each realization we compared the performance of these policies: we measured the mean reward difference betwen PPO and CEsc, and between PPO+GP and CEsc. The plot represents the distribution of reward differences observed at a given noise strength---we plot the mean and the standard deviation of the mean reward differences observed. In an equation, we plot the means $\mathbb{E}_{P}[\mu_P^{\text{DRL}}-\mu_P^{\text{CEsc}}]$, where $P$ are the paramter values, $\mu_P^{\text{DRL}}$ is the mean reward for a DRL policy trained on the problem with parameter values $P$, and, similarly, $\mu_P^{\text{CEsc}}$ is the mean reward of the optimal constant escpament policy for parameter values $P$.}
  \label{fig:stability}
\end{figure}

Our main result in this respect is Fig. \ref{fig:stability}. There, we
plot the \emph{average episode reward difference} between the two
DRL-based methods we considered, and the optimal CEsc strategy. This
figure shows that, for a wide range of parameter values, DRL-based
strategies can have a considerable advantage over an optimized CEsc
policy (the single-species optimal solution).

\hypertarget{discussion}{%
\section{Discussion}\label{discussion}}

Fisheries are complex ecosystems, with interactions between species
leading to highly non-linear dynamics. While current models for
population estimation include many important aspects of this complexity,
it is still common to use simplified dynamical models in order to guide
decision making. This provides easily interpretable solutions, such as
CMort policies. There is a drawback here, however: due to the simplicity
of these dynamical models, the policies might not respond effectively in
many situations---situations where the predictions of these simple
models deviate considerably from reality. Because of this, policies such
as MSY have faced pushback and are believed to have contributed to the
depletion of fish stocks (Worm et al. 2006; Costello et al. 2016).

We propose an alternative approach to the problem of fishery control: to
use a more expressive -- albeit more complicated -- dynamical model to
guide decision making. Furthermore, rather than computing the optimal
control policy for the model (something that is impossible in practice
for complex systems), we use deep reinforcement learning to obtain a
``pretty darn good'' policy. This policy is estimated in a
\emph{model-free} setting, i.e., the agent treats the dynamical model
(e.g.~eq. \eqref{eq:3d model}) as a black box of input-output pairs. By
not relying on precise knowledge of the model's parameter values, but
rather just relying on input-output statistics, model-free approaches
have gained traction in a variety of control theory settings (see, e.g.,
(Sato 2019; Ramirez, Yu, and Perrusquia 2022; Zeng et al. 2019)).

We compare deep reinforcement learning-based policies against classical
management strategies (CMort and CEsc). While the latter are inspired by
the shape of optimal solutions in the single-species setting, they are
optimized in a model-free way as well: e.g.~the optimal mortality rate
is computed empirically from simulated data.

Because of the simplicity of the classical policy functions, the optimal
such policy may be easily estimated through a grid-search. This is the
case since these policy functions are only specified by one or two
parameters (respectively in the single fishery, or two fishery cases).
In contrast, DRL optimizes over the more expressive -- and more
complicated -- family of policies parametrized by a neural network.
Neural networks are often used as flexible function approximators that
can be efficiently optimized over.

We showed that for sufficiently complex management scenarios -- Models 3
and 4 -- DRL-based management strategies perform significantly better
than CEsc. This with respect to both average rewards received and
conservation goals. In this sense, an approximate solution to a more
complicated and expressive model, can outperform the optimal solution of
the single-species problem---even when the parameters of the
single-species solution are empirically optimized.

We found that the optimal CMort policy surprisingly performs much better
than CEsc (Fig. \ref{fig:main}). However, it can be observed in Fig.
\ref{fig:fracmsy} that the CMort strategy faces a trade-off: high
sustainability is achieved for sub-optimal mortality rates, but only at
a significant decrease in the mean episode reward. We expect that with
increasing ecosystem complexity this phenomenon might become more
pronounced. We can understand this as a consequence of the rigidity of
classical strategies: the simplicity of their expressions, depending
only on a few parameters, means that policy optimization is constrained
to a rather reduced subset of the space of possible policies.

The question of when multi-species models are well-approximated by
single-species models was studied in detail in (Burgess et al. 2017).
Here our approach is dual to that of the aforementioned paper. Rather
than first optimizing a single-species model to approximate a more
complex model and then finding the MSY value for the single-species
model, we used simulated data to optimize CMort and CEsc directly on the
three-species model. We do not investigate further whether our
three-species models are well approximated by a single-species model in
the sense of (Burgess et al. 2017). However: \emph{1)} Because the
interaction terms in \eqref{eq:3d model} are about an order of magnitude
smaller than \(r_X\) and \(r_Y\), Models 2-4 are ``close'' in
parameter-space to a single-species model. \emph{2)} The fact that for
Model 2 all strategies match in performance suggests that
\eqref{eq:3d model} might be well-approximated by a single species
model. This in turn suggests that the reason that DRL outperforms both
single-species strategies for Models 3 and 4 is not due a lack of a
single-species approximation for either \(X\) or \(Y\), but due to the
complexity of having two harvested species. Moreover, the
non-stationarity in Model 4 maintained the advantage of DRL over CEsc
and CMort. Here, one may have expected an exacerbation of that advantage
due to non-stationarities introducing biases to single-species
approximate models (Burgess et al. 2017). We did, however, measure a
decrease in the sustainability of CEsc in Model 4 with respect to Model
3.

Finally, we performed a stability analysis to ensure that the advantage
of DRL-based techniques over CEsc is a ubiquitous phenomenon and not a
result of a lucky selection of parameter values. We found (in Fig.
\ref{fig:stability}) that an advantage can be observed even for
relatively high-noise perturbations of the parameters---noise with a
variance of 20\% of the parameter values. The aforementioned Figure
summarizes the statistics of 100 parameter perturbations so that we may
expect perturbations of up to 60\% (i.e.~three sigmas) in the parameter
values to appear in these statistics.

\hypertarget{future-directions}{%
\subsection{Future directions}\label{future-directions}}

There are a number of interesting directions that would be interseting
to explore in future work.

First, benchmarking our results against increasingly more compex and
realistic fishery models. This would include non-stationarities in the
dynamical parameters to accurately reflect the effects of climate change
on the ecosystem. This added complexity would likely pose a
computational challenge---in future work we will likely need to test
several different DRL training algorithms (see e.g. (Lapeyrolerie et al.
2022) for a non-exhaustive list), and it is very likely that
hyperparameter tuning will need to be performed. Moreover, it may be
that larger neural networks than the one we used in this research will
be needed for the policy function. This all will mean that considerable
technical work will be needed in order to make this next step
computationally feasible (e.g.~we might need to make more extensive use
of GPUs and parallelization).

Second, to account for noisy estimates of the system's state and
imperfect policy implementation. This could be done straightforwardly,
albeit it might increase the training time before DRL approaches
converge, as well as introducing the need for hyperparameter tuning.

Third, to account for the systematic uncertainties behind the dynamics
of the ecosystem---that is, to account for model biases with respect to
reality. Here, one can employ tools from curriculum learning in order to
train an agent that is \emph{generally capable} of good management over
a range of different dynamical models. This way, one can incorporate
different models---expressing different aspects of the ecosystem---into
the learning process of the agent. We believe that this step will likely
be necessary if DRL algorithms are to be applied successfully in the
fishery management problem. Curriculum learning is rather expensive
computationally, however, and involves a non-trivial \emph{curriculum
design} which will guide the agent in its learning process. This way,
considerable technical work would be needed for this direction.

\hypertarget{acknowledgements}{%
\subsection{Acknowledgements}\label{acknowledgements}}

The title of this piece references a mathematical biology workshop at
NIMBioS organized by Paul Armsworth, Alan Hastings, Megan Donahue, and
Carl Towes in 2011 which first sought to emphasize `pretty darn good'
control solutions to more realistic problems over optimal control to
idealized ones. This material is based upon work supported by the
National Science Foundation under Grant No.~DBI-1942280.

\appendix

\hypertarget{appendix-results-for-stationary-models}{%
\section{Appendix: Results for stationary
models}\label{appendix-results-for-stationary-models}}

\begin{figure}
\includegraphics[width=6in]{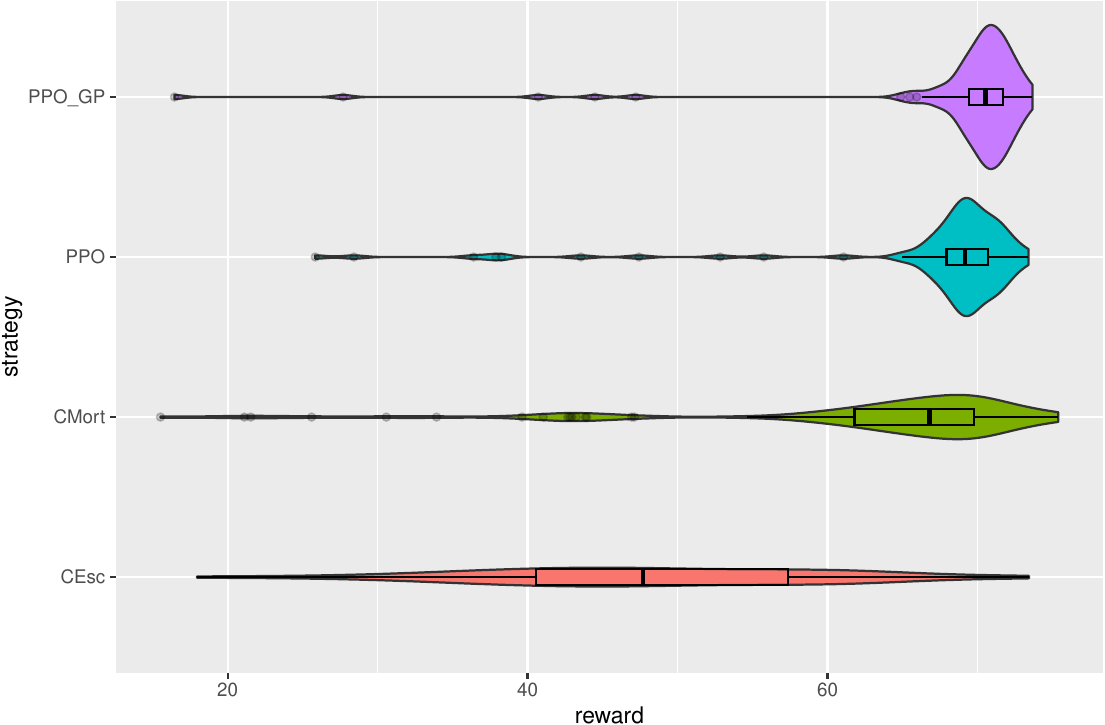} \caption{Reward distributions for the four strategies considered. These are based on 100 evaluation episodes of Model 3 in Table 1. We denote CEsc for constant escapement, CMort for constant mortality, PPO for the output policy of the PPO optimization algorithm, and PPO GP for the Gaussian process interpolation of the PPO policy.}\label{fig:main_32}
\end{figure}

\begin{figure}
\includegraphics[width=6in]{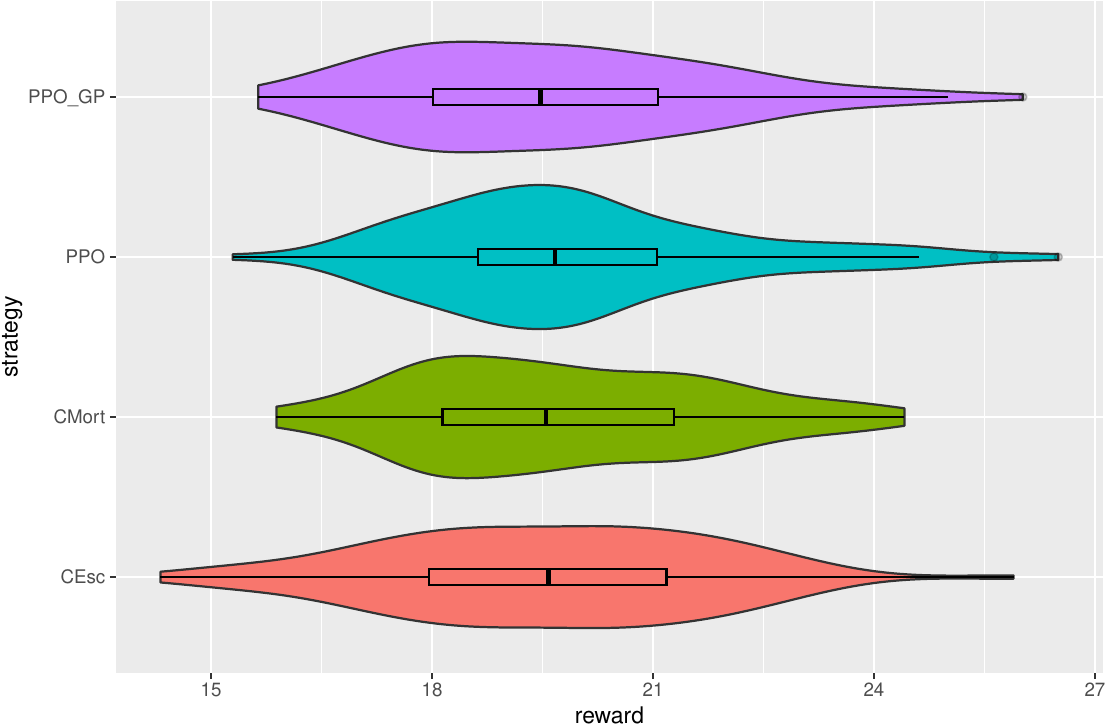} \caption{Reward distributions for the four strategies considered. These are based on 100 evaluation episodes of Model 2 in Table 1. We denote CEsc for constant escapement, CMort for constant mortality, PPO for the output policy of the PPO optimization algorithm, and PPO GP for the Gaussian process interpolation of the PPO policy.}\label{fig:main_31}
\end{figure}

\begin{figure}
\includegraphics[width=6in]{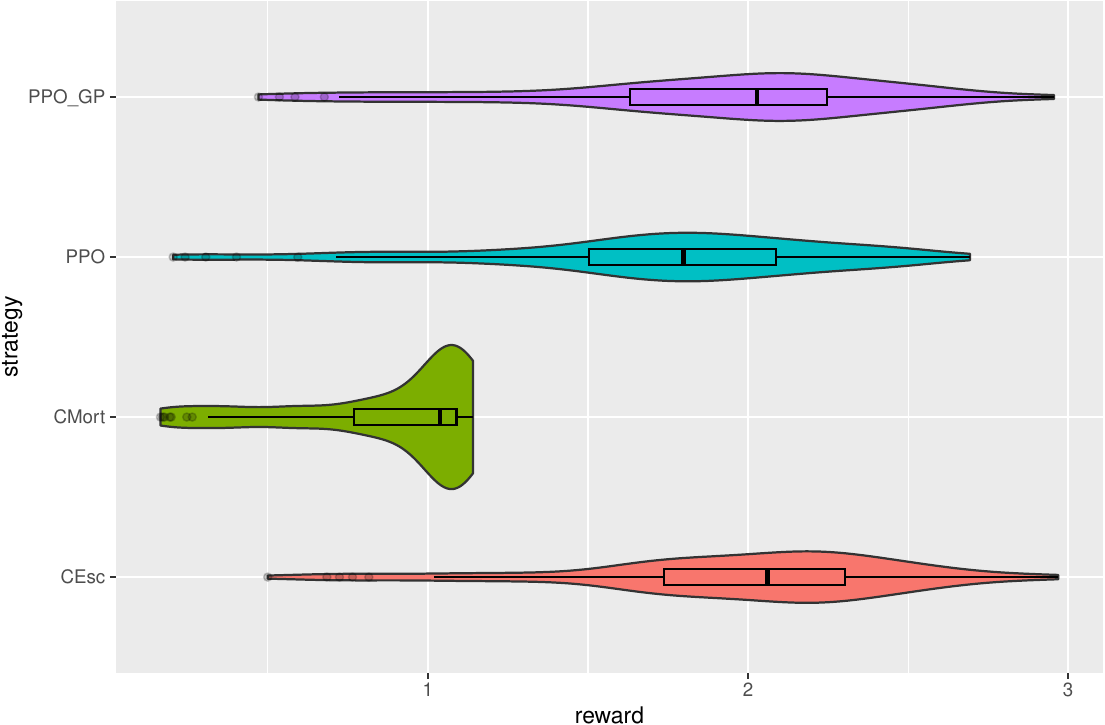} \caption{Reward distributions for the four strategies considered. These are based on 100 evaluation episodes of Model 1 in Table 1. We denote CEsc for constant escapement, CMort for constant mortality, PPO for the output policy of the PPO optimization algorithm, and PPO GP for the Gaussian process interpolation of the PPO policy.}\label{fig:main_11}
\end{figure}

In the main text we focused on the non-stationary model (``three
species, two fisheries, non-stationary'' in Table 1) for the sake of
space and because our results were most compelling there. Here we
present the reward distributions for the other models considered---the
three stationary models, lines 1-3 in Table 1. These results are shown
in Figs. \ref{fig:main_32}, \ref{fig:main_31} and \ref{fig:main_11}.

\hypertarget{appendix-ppo-policy-function-for-non-stationary-model}{%
\section{Appendix: PPO policy function for non-stationary
model}\label{appendix-ppo-policy-function-for-non-stationary-model}}

\begin{figure}
  \centering
  \includegraphics[scale=0.35]{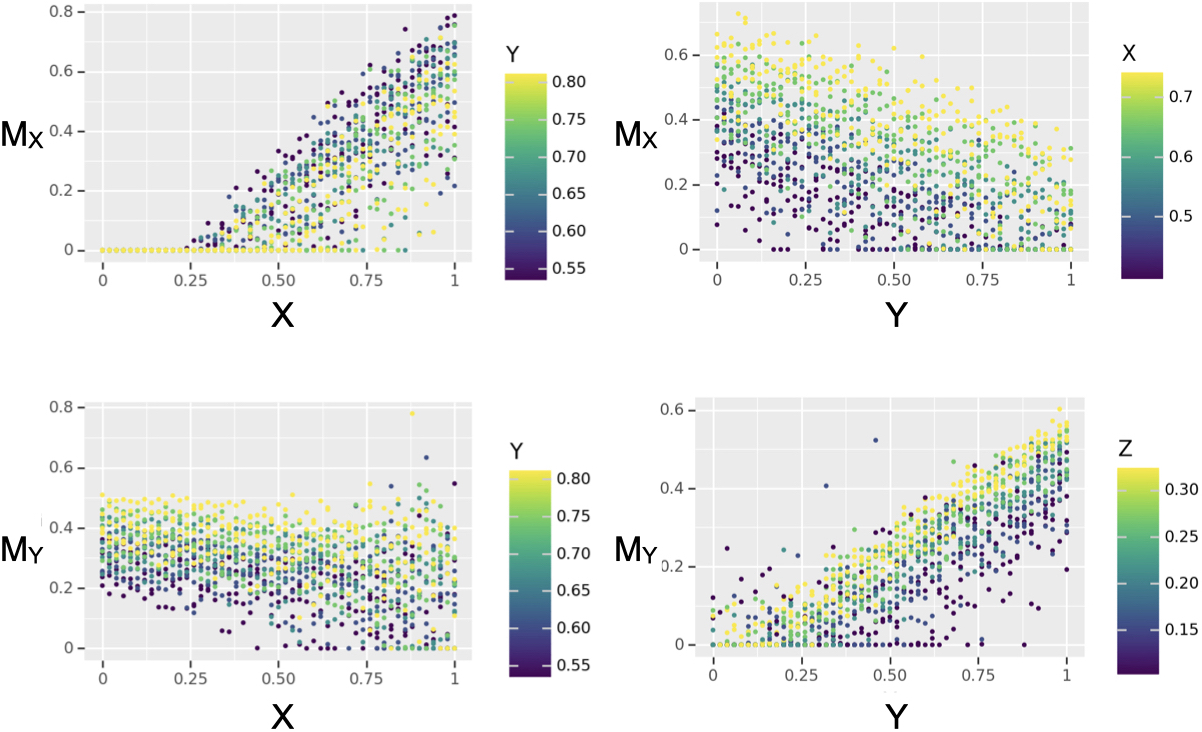}
  \caption{Plots of the PPO policy $\pi_{\text{PPO}}$ along several relevant axes. Here $M_X$ and $M_Y$ are the $X$ and $Y$ components of the policy function. The values of the plots are generated in the following way: For each variable $X$, $Y$, and $Z$, the time-series of evaluation episodes are used to generate a window of typical values that variable attains when controlled by $\pi_{\text{PPO}}$. Then, for each plot either $X$ or $Y$ was evaluated on 100 values in $[0,1]$ along the $x$ axis, while the other variables (resp. $Y$ and $Z$, or $X$ and $Z$) were varied within the typical window computed before. Within this window, 5 values are used. The value of one of the latter two variables were visualized as color. This scatter data was used as an input to generate $\pi_{\text{PPO+GP}}$, visualized in the main text.}
  \label{fig:ppo-policy}
\end{figure}

In the main text, Fig. \ref{fig:gpp-policy}, we presented a
visualization of the PPO+GP policy function obtained for the ``three
species, two fisheries, non-stationary'' model. This policy function is
a Gaussian process regression of scatter data of the PPO policy
function. In Fig. \ref{fig:ppo-policy} we present a representation of
this scatter data in a similar format as Fig. \ref{fig:gpp-policy}.

\hypertarget{appendix-gaussian-process-interpolation}{%
\section{Appendix: Gaussian process
interpolation}\label{appendix-gaussian-process-interpolation}}

Here we summarize the procedure used to interpolate the PPO policy
(visualized in Fig. \ref{fig:ppo-policy}). We use the
\emph{GaussianProcessRegressor} object of the \emph{sklearn} Python
library with a kernel given by \begin{align*}
  \text{RBF}(\text{length scale = 10}) + \text{WhiteNoise}(\text{noise level = 0.1}).
\end{align*} This interpolation method is applied to scatter data of the
PPO policy evaluated on 3 different grids on \((X,Y,Z)\) states:
\(G_X\), a \(51\times 5 \times 5\) grid; \(G_Y\), a
\(5\times 51 \times 5\) grid; and \(G_Z\), a \(5\times 5 \times 51\)
grid. This combination of grids was used instead of a single dense grid
in order to reduce the computational intensity of the interpolation
procedure. For \(G_X\), the 5 values for \(Y\) and \(Z\) were varied in
a ``popular window,'' i.e.~episode time-series data was used to
determine windows of \(Y\) and \(Z\) values which were most likely. The
grids \(G_Y\) and \(G_Z\) were generated in a similar fashion,
\emph{mutatis mutandis}.\footnote{
The raw dataset is found at the data/results\_data/2FISHERY/RXDRIFT sub-directory in the repository with the source code and data linked above.
Scatter plots visualizing this policy are shown in App. B.
} The length scale and noise level values of this kernel were chosen
arbitrarily---no hyperparameter tuning was needed to produce
satisfactory interpolation, as will be shown in the results section.

\hypertarget{references}{%
\section*{References}\label{references}}
\addcontentsline{toc}{section}{References}

\hypertarget{refs}{}
\begin{CSLReferences}{1}{0}
\leavevmode\vadjust pre{\hypertarget{ref-anderson2007optimal}{}}%
Anderson, Brian DO, and John B Moore. 2007. \emph{Optimal Control:
Linear Quadratic Methods}. Courier Corporation.

\leavevmode\vadjust pre{\hypertarget{ref-atari1}{}}%
Bellemare, Marc G, Yavar Naddaf, Joel Veness, and Michael Bowling. 2013.
{``The Arcade Learning Environment: An Evaluation Platform for General
Agents.''} \emph{Journal of Artificial Intelligence Research} 47:
253--79.

\leavevmode\vadjust pre{\hypertarget{ref-burgess2017describing}{}}%
Burgess, Matthew G, Henrique C Giacomini, Cody S Szuwalski, Christopher
Costello, and Steven D Gaines. 2017. {``Describing Ecosystem Contexts
with Single-Species Models: A Theoretical Synthesis for Fisheries.''}
\emph{Fish and Fisheries} 18 (2): 264--84.

\leavevmode\vadjust pre{\hypertarget{ref-millie-3}{}}%
Chapman, Melissa, Lily Xu, Marcus Lapeyrolerie, and Carl Boettiger.
2023. {``Bridging Adaptive Management and Reinforcement Learning for
More Robust Decisions.''} \emph{Philosophical Transactions of the Royal
Society B (Accepted)}.

\leavevmode\vadjust pre{\hypertarget{ref-Clark1990}{}}%
Clark, Colin W. 1990. \emph{{Mathematical Bioeconomics: The Optimal
Management of Renewable Resources, 2nd Edition}}. Wiley-Interscience.

\leavevmode\vadjust pre{\hypertarget{ref-Clark1973}{}}%
Clark, Colin W. 1973. {``Profit Maximization and the Extinction of
Animal Species.''} \emph{Journal of Political Economy} 81 (4): 950--61.
\url{https://doi.org/10.1086/260090}.

\leavevmode\vadjust pre{\hypertarget{ref-collins2001internal}{}}%
Collins, MSFB, SFB Tett, and C Cooper. 2001. {``The Internal Climate
Variability of HadCM3, a Version of the Hadley Centre Coupled Model
Without Flux Adjustments.''} \emph{Climate Dynamics} 17: 61--81.

\leavevmode\vadjust pre{\hypertarget{ref-Costello2016}{}}%
Costello, Christopher, Daniel Ovando, Tyler Clavelle, C. Kent Strauss,
Ray Hilborn, Michael C. Melnychuk, Trevor A Branch, et al. 2016.
{``{Global fishery prospects under contrasting management regimes}.''}
\emph{Proceedings of the National Academy of Sciences} 113 (18):
5125--29. \url{https://doi.org/10.1073/pnas.1520420113}.

\leavevmode\vadjust pre{\hypertarget{ref-fusion1}{}}%
Degrave, Jonas, Federico Felici, Jonas Buchli, Michael Neunert, Brendan
Tracey, Francesco Carpanese, Timo Ewalds, et al. 2022. {``Magnetic
Control of Tokamak Plasmas Through Deep Reinforcement Learning.''}
\emph{Nature} 602 (7897): 414--19.

\leavevmode\vadjust pre{\hypertarget{ref-franccois2018introduction}{}}%
François-Lavet, Vincent, Peter Henderson, Riashat Islam, Marc G
Bellemare, Joelle Pineau, et al. 2018. {``An Introduction to Deep
Reinforcement Learning.''} \emph{Foundations and Trends{\textregistered}
in Machine Learning} 11 (3-4): 219--354.

\leavevmode\vadjust pre{\hypertarget{ref-gordon2000simulation}{}}%
Gordon, Chris, Claire Cooper, Catherine A Senior, Helene Banks, Jonathan
M Gregory, Timonthy C Johns, John FB Mitchell, and Richard A Wood. 2000.
{``The Simulation of SST, Sea Ice Extents and Ocean Heat Transports in a
Version of the Hadley Centre Coupled Model Without Flux Adjustments.''}
\emph{Climate Dynamics} 16: 147--68.

\leavevmode\vadjust pre{\hypertarget{ref-Gordon1954}{}}%
Gordon, H. Scott, and Chicago Press. 1954. {``{The Economic Theory of a
Common-Property Resource: The Fishery}.''} \emph{Journal of Political
Economy} 62 (2): 124--42. \url{https://doi.org/10.1086/257497}.

\leavevmode\vadjust pre{\hypertarget{ref-mbpo}{}}%
Janner, Michael, Justin Fu, Marvin Zhang, and Sergey Levine. 2019.
{``When to {Trust} {Your} {Model}: {Model}-{Based} {Policy}
{Optimization}.''} \emph{arXiv:1906.08253 {[}Cs, Stat{]}}, November.
\url{http://arxiv.org/abs/1906.08253}.

\leavevmode\vadjust pre{\hypertarget{ref-rl-intro}{}}%
Lapeyrolerie, Marcus, Melissa S Chapman, Kari EA Norman, and Carl
Boettiger. 2022. {``Deep Reinforcement Learning for Conservation
Decisions.''} \emph{Methods in Ecology and Evolution} 13 (11): 2649--62.

\leavevmode\vadjust pre{\hypertarget{ref-mangel2006theoretical}{}}%
Mangel, Marc. 2006. \emph{The Theoretical Biologist's Toolbox:
Quantitative Methods for Ecology and Evolutionary Biology}. Cambridge
University Press.

\leavevmode\vadjust pre{\hypertarget{ref-Marescot2013}{}}%
Marescot, Lucile, Guillaume Chapron, Iadine Chadès, Paul L. Fackler,
Christophe Duchamp, Eric Marboutin, and Olivier Gimenez. 2013.
{``Complex Decisions Made Simple: A Primer on Stochastic Dynamic
Programming.''} \emph{Methods in Ecology and Evolution} 4 (9): 872--84.
\url{https://doi.org/10.1111/2041-210X.12082}.

\leavevmode\vadjust pre{\hypertarget{ref-may77}{}}%
May, Robert M. 1977. {``Thresholds and Breakpoints in Ecosystems with a
Multiplicity of Stable States.''} \emph{Nature} 269 (5628): 471--77.

\leavevmode\vadjust pre{\hypertarget{ref-atari2}{}}%
Mnih, Volodymyr, Koray Kavukcuoglu, David Silver, Alex Graves, Ioannis
Antonoglou, Daan Wierstra, and Martin Riedmiller. 2013. {``Playing Atari
with Deep Reinforcement Learning.''} \emph{arXiv Preprint
arXiv:1312.5602}.

\leavevmode\vadjust pre{\hypertarget{ref-moerland2023model}{}}%
Moerland, Thomas M, Joost Broekens, Aske Plaat, Catholijn M Jonker, et
al. 2023. {``Model-Based Reinforcement Learning: A Survey.''}
\emph{Foundations and Trends{\textregistered} in Machine Learning} 16
(1): 1--118.

\leavevmode\vadjust pre{\hypertarget{ref-chatgpt}{}}%
OpenAI. 2022. {``{C}hat{G}{P}{T}: {O}ptimizing {L}anguage {M}odels for
{D}ialogue.''} \url{https://openai.com/blog/chatgpt/}.

\leavevmode\vadjust pre{\hypertarget{ref-polydoros2017survey}{}}%
Polydoros, Athanasios S, and Lazaros Nalpantidis. 2017. {``Survey of
Model-Based Reinforcement Learning: Applications on Robotics.''}
\emph{Journal of Intelligent \& Robotic Systems} 86 (2): 153--73.

\leavevmode\vadjust pre{\hypertarget{ref-pope2000impact}{}}%
Pope, VD, ML Gallani, PR Rowntree, and RA Stratton. 2000. {``The Impact
of New Physical Parametrizations in the Hadley Centre Climate Model:
HadAM3.''} \emph{Climate Dynamics} 16: 123--46.

\leavevmode\vadjust pre{\hypertarget{ref-Punt2016}{}}%
Punt, André E, Doug S Butterworth, Carryn L de Moor, José A A De
Oliveira, and Malcolm Haddon. 2016. {``{Management strategy evaluation:
best practices}.''} \emph{Fish and Fisheries} 17 (2): 303--34.
\url{https://doi.org/10.1111/faf.12104}.

\leavevmode\vadjust pre{\hypertarget{ref-RAMLegacyDB}{}}%
RAM Legacy Stock Assessment Database. 2020. {``RAM Legacy Stock
Assessment Database V4.491.''}
\url{https://doi.org/10.5281/zenodo.3676088}.

\leavevmode\vadjust pre{\hypertarget{ref-ramirez2022}{}}%
Ramirez, Jorge, Wen Yu, and Adolfo Perrusquia. 2022. {``Model-Free
Reinforcement Learning from Expert Demonstrations: A Survey.''}
\emph{Artificial Intelligence Review}, 1--29.

\leavevmode\vadjust pre{\hypertarget{ref-riahi2017shared}{}}%
Riahi, Keywan, Detlef P Van Vuuren, Elmar Kriegler, Jae Edmonds, Brian C
O'neill, Shinichiro Fujimori, Nico Bauer, et al. 2017. {``The Shared
Socioeconomic Pathways and Their Energy, Land Use, and Greenhouse Gas
Emissions Implications: An Overview.''} \emph{Global Environmental
Change} 42: 153--68.

\leavevmode\vadjust pre{\hypertarget{ref-sato2019}{}}%
Sato, Yoshiharu. 2019. {``Model-Free Reinforcement Learning for
Financial Portfolios: A Brief Survey.''} \emph{arXiv Preprint
arXiv:1904.04973}.

\leavevmode\vadjust pre{\hypertarget{ref-Schaefer1954}{}}%
Schaefer, Milner B. 1954. {``{Some aspects of the dynamics of
populations important to the management of the commercial marine
fisheries}.''} \emph{Bulletin of the Inter-American Tropical Tuna
Commission} 1 (2): 27--56. \url{https://doi.org/10.1007/BF02464432}.

\leavevmode\vadjust pre{\hypertarget{ref-fusion2}{}}%
Seo, J, Y-S Na, B Kim, CY Lee, MS Park, SJ Park, and YH Lee. 2022.
{``Development of an Operation Trajectory Design Algorithm for Control
of Multiple 0D Parameters Using Deep Reinforcement Learning in KSTAR.''}
\emph{Nuclear Fusion} 62 (8): 086049.

\leavevmode\vadjust pre{\hypertarget{ref-sethi2019optimal}{}}%
Sethi, Suresh P, and Suresh P Sethi. 2019. \emph{What Is Optimal Control
Theory?} Springer.

\leavevmode\vadjust pre{\hypertarget{ref-Worm2006}{}}%
Worm, Boris, Edward B Barbier, Nicola Beaumont, J Emmett Duffy, Carl
Folke, Benjamin S Halpern, Jeremy B C Jackson, et al. 2006. {``{Impacts
of biodiversity loss on ocean ecosystem services.}''} \emph{Science (New
York, N.Y.)} 314 (5800): 787--90.
\url{https://doi.org/10.1126/science.1132294}.

\leavevmode\vadjust pre{\hypertarget{ref-zeng2019}{}}%
Zeng, Deze, Lin Gu, Shengli Pan, Jingjing Cai, and Song Guo. 2019.
{``Resource Management at the Network Edge: A Deep Reinforcement
Learning Approach.''} \emph{IEEE Network} 33 (3): 26--33.

\leavevmode\vadjust pre{\hypertarget{ref-zhang2019near}{}}%
Zhang, Yinyan, Shuai Li, and Liefa Liao. 2019. {``Near-Optimal Control
of Nonlinear Dynamical Systems: A Brief Survey.''} \emph{Annual Reviews
in Control} 47: 71--80.

\end{CSLReferences}

\bibliographystyle{spphys}
\bibliography{bibliography.bib}

\end{document}